\documentclass[10pt,twocolumn,letterpaper]{article}

\PassOptionsToPackage{dvipsnames,table}{xcolor}

\newcommand{\ie}{\textit{i}.\textit{e}.}
\newcommand{\cf}{\textit{cf.}}

\newcommand{\vs}{\textit{vs}.}

\usepackage[pagenumbers]{cvpr}   

\usepackage{needspace}
\usepackage{algorithm}
\usepackage{algorithmic}
\usepackage{amsmath}
\usepackage{amssymb}
\usepackage{bm}
\usepackage{booktabs}       
\usepackage{multirow}       
\usepackage[table]{xcolor}  

\definecolor{figgreen}{HTML}{78985D}
\definecolor{cellblue}{RGB}{205, 215, 255}   
\definecolor{diffgray}{RGB}{140, 140, 140}   
\definecolor{difforange}{RGB}{225, 125, 0}   
\definecolor{citeblue}{RGB}{24, 110, 210}    

\definecolor{cvprblue}{rgb}{0.21,0.49,0.74}
\usepackage[breaklinks,colorlinks,allcolors=cvprblue]{hyperref}
\usepackage{float}
\definecolor{figpurple}{HTML}{A37DB7}

\usepackage{tcolorbox}
\usepackage{fancyhdr}
\IfFileExists{microtype.sty}{\usepackage{microtype}}{}

\definecolor{hkustblue}{HTML}{003366}
\definecolor{hkustgold}{HTML}{996600}
\definecolor{longL}{HTML}{0085FA}
\definecolor{longO}{HTML}{FF5845}
\definecolor{longN}{HTML}{14804F}
\definecolor{longG}{HTML}{2E5FD8}

\newcommand{\LONGGroupBanner}{%
  \href{https://long-group.cse.ust.hk/}{%
    {\fontsize{10.5pt}{12pt}\selectfont\rmfamily
     {\bfseries\textcolor{longL}{L}\textcolor{longO}{O}\textcolor{longN}{N}\textcolor{longG}{G}}%
     \hspace{0.25em}\textcolor{black!82}{\bfseries Group}}%
  }%
}
\newcommand{\LONGGroupHeader}{%
  {\bfseries\textcolor{longL}{L}\textcolor{longO}{O}\textcolor{longN}{N}\textcolor{longG}{G}}~Group%
}

\newcommand\blfootnote[1]{%
  \begingroup
  \renewcommand\thefootnote{}\footnote{#1}%
  \addtocounter{footnote}{-1}%
  \endgroup
}

\newcommand{\shorttitle}{Dual-Manifold Variational Autoencoders}
\newcommand{\arxivlabel}{Research Paper}
\newcommand{\projectlinks}{%
  \textbf{Project Page:}~\href{https://ustcchx.github.io/DM-VAE-page/}{\texttt{ustcchx.github.io/DM-VAE-page}}
}

\makeatletter
\def\@maketitle{%
  \newpage
  \null
  \thispagestyle{plain}%
  \vspace*{-18pt}%
  \noindent\begin{minipage}{\textwidth}
    \IfFileExists{hkust_logo.pdf}{%
      \href{https://hkust.edu.hk/}{\raisebox{-0.24\height}{\includegraphics[height=0.56cm]{hkust_logo.pdf}}}%
    }{%
      \IfFileExists{hkust_logo.png}{%
        \href{https://hkust.edu.hk/}{\raisebox{-0.24\height}{\includegraphics[height=0.56cm]{hkust_logo.png}}}%
      }{%
        \href{https://hkust.edu.hk/}{%
          {\color{hkustgold}\rule[-0.14cm]{2.2pt}{0.46cm}}\hspace{4pt}%
          {\fontsize{10.8pt}{12pt}\selectfont\bfseries\color{hkustblue}HKUST}%
        }%
      }%
    }%
    \hspace{8pt}{\color{black!22}\rule[-0.16cm]{0.6pt}{0.48cm}}\hspace{8pt}%
    \LONGGroupBanner
    \hfill
    {\footnotesize\scshape\color{black!55}\arxivlabel}%
    \par\vspace{5pt}%
    {\color{hkustblue!30}\hrule height 0.65pt}%
  \end{minipage}%
  \vspace{13pt}%
  \begin{center}%
    {\LARGE\bfseries\color{hkustblue!30!black} \@title \par}%
    \vspace{10pt}%
    {\large\lineskip .45em
      \begin{tabular}[t]{c}%
        \@author
      \end{tabular}\par}%
    \vspace{6pt}%
    {\small \projectlinks \par}%
    \vspace{9pt}%
  \end{center}%
}

\renewenvironment{abstract}{%
  \thispagestyle{plain}%
  \begin{tcolorbox}[
    colback=hkustblue!3,
    colframe=hkustblue!16,
    boxrule=0.5pt,
    arc=3pt,
    boxsep=0pt,
    left=8.5pt,
    right=8.5pt,
    top=6.5pt,
    bottom=6.5pt
  ]
  {\centering\normalsize\bfseries\color{hkustblue!85!black} Abstract\par}%
  \vspace{3.5pt}\noindent\small\normalfont\ignorespaces
}{%
  \end{tcolorbox}%
  \vspace{4pt}%
}

\renewcommand{\headrulewidth}{0.4pt}
\renewcommand{\headrule}{\hbox to\headwidth{\color{black!16}\leaders\hrule height \headrulewidth\hfill}}
\fancypagestyle{plain}{%
  \fancyhf{}%
  \fancyfoot[C]{\small\color{black!65}\thepage}%
  \renewcommand{\headrulewidth}{0pt}%
}

\renewcommand{\maketitlesupplementary}{%
  \clearpage
  \twocolumn[{%
    \centering
    \vspace*{2pt}
    {\Large\bfseries\color{hkustblue!30!black} Appendix: \thetitle \par}
    \vspace{7pt}
    {\color{hkustblue!25}\hrule height 0.6pt}
    \vspace{14pt}
  }]%
}
\makeatother

\def\confName{CVPR}
\def\confYear{2027}

\title{Dual-Manifold Variational Autoencoders}

\author{
\href{https://ustcchx.github.io/}{\textcolor{black}{Hongxu Chen}}$^{*}$, \quad
\href{https://chunwang.site/}{\textcolor{black}{Chun Wang}}$^{*}$, \quad
Yanghao Wang, \\
Hongxiang Li, \quad
Zhen Wang, \quad
Zhixiang Zhang, \quad
\href{https://zjuchenlong.github.io/}{\textcolor{black}{Long Chen}}$^{\dagger}$ \\[0.4em]
{\normalsize The Hong Kong University of Science and Technology}
}

\begin{document}
\maketitle
\blfootnote{$^{*}$Equal contribution. $^{\dagger}$Corresponding author (\texttt{longchen@ust.hk}).}

\begin{abstract}
In latent diffusion models, expanding the latent dimension of variational autoencoders (VAEs) is essential to push the upper bound of reconstruction quality. However, without explicit structural constraints, such unregularized expansion easily yields a rugged and ill-conditioned latent manifold, causing generated latents to drift off the manifold during generation (\ie, \textbf{off-manifold} problem). Recent works smooth this manifold by using external foundation models (\eg, DINOv2) as a topological regularizer. However, we argue that tying the latent space to frozen external priors inherently restricts generalization: bound by their pre-training distributions and semantic biases, they may fail to capture faithful visual details on out-of-distribution domains. Observing that low-dimensional latents naturally form a smoother, generation-friendly manifold, we propose \textbf{dual-manifold VAE (DM-VAE)} to leverage internal representations as a topological regularizer. Specifically, DM-VAE pairs a high-dimensional reconstructive branch with a low-dimensional branch that extracts a clean topological skeleton, anchoring the high-dimensional manifold to this smooth subspace. Building upon this framework, we further introduce topological representation alignment (TRA) to supervise intermediate generator features during training, and internal guidance with topological skeleton ($\text{IG}_{TS}$) to guide structural components while preserving high-frequency details at inference. On ImageNet-1K ($256\times256$), our foundation-model-free DM-VAE achieves a state-of-the-art rFID of 0.12 and a generation $\text{FID}_{50K}$ of 1.28 (w/o CFG) and 1.20 (w/ CFG), and further reaches 1.11 when orthogonally combined with DINOv2.
\end{abstract}
\section{Introduction}
\label{sec:intro}

\begin{figure}
    \centering
    \includegraphics[width=0.96\linewidth]{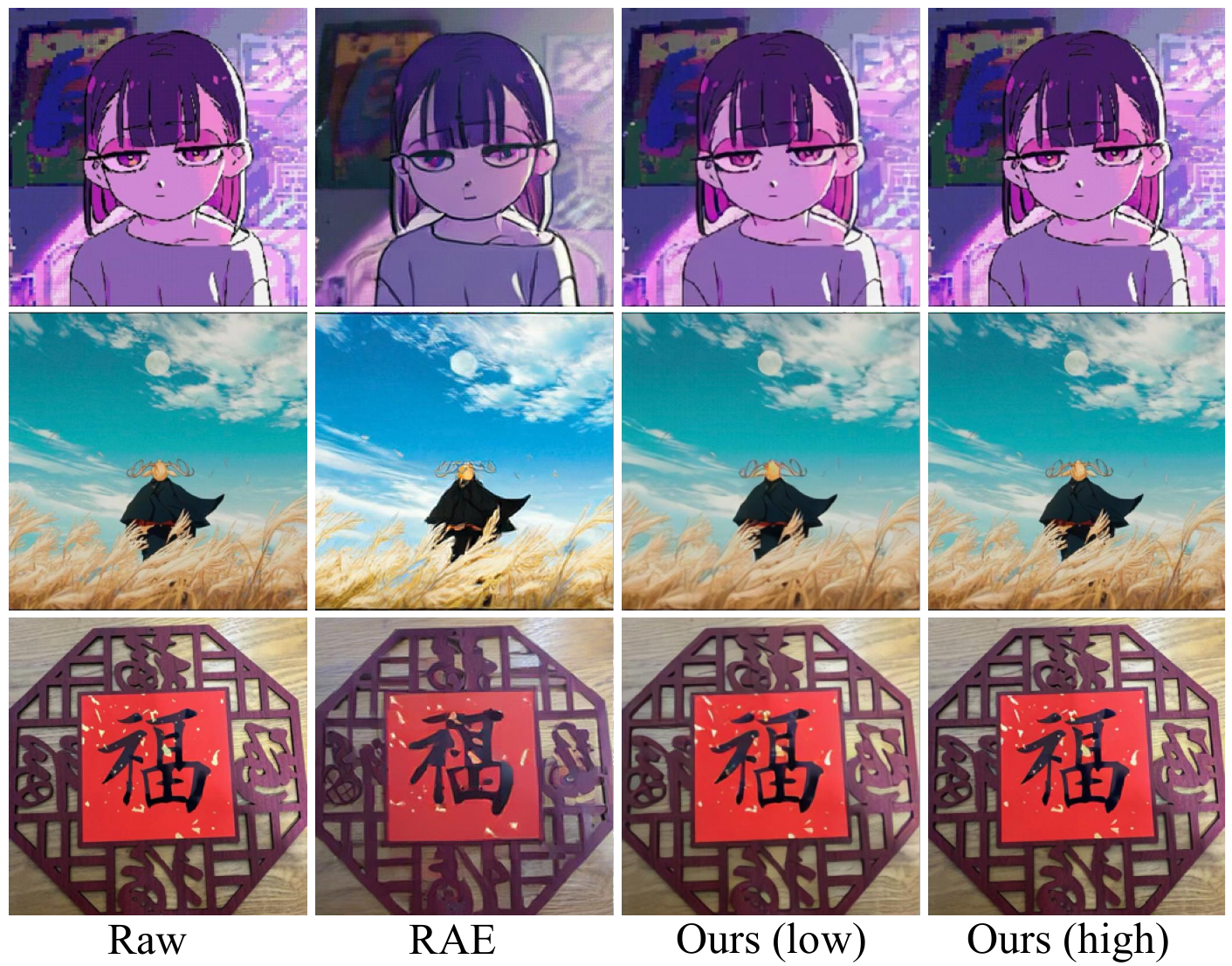}
    \vspace{-1em}
    \caption{\textbf{Qualitative comparison of out-of-distribution image reconstruction.} All methods are trained on ImageNet-1K. \textbf{Without relying on external foundation models}, our DM-VAE (\texttt{f16d64}) exhibits superior generalization. Average PSNR~\citep{hore2010image} on above three samples: RAE~\citep{zheng2026diffusion} (16.32~dB), Ours (low branch) (27.19~dB), and Ours (high branch) (\textbf{30.10~dB}).}
    \label{fig:1}
    \vspace{-1.5em}
\end{figure}

\begin{figure*}
    \centering
\includegraphics[width=0.96\linewidth]{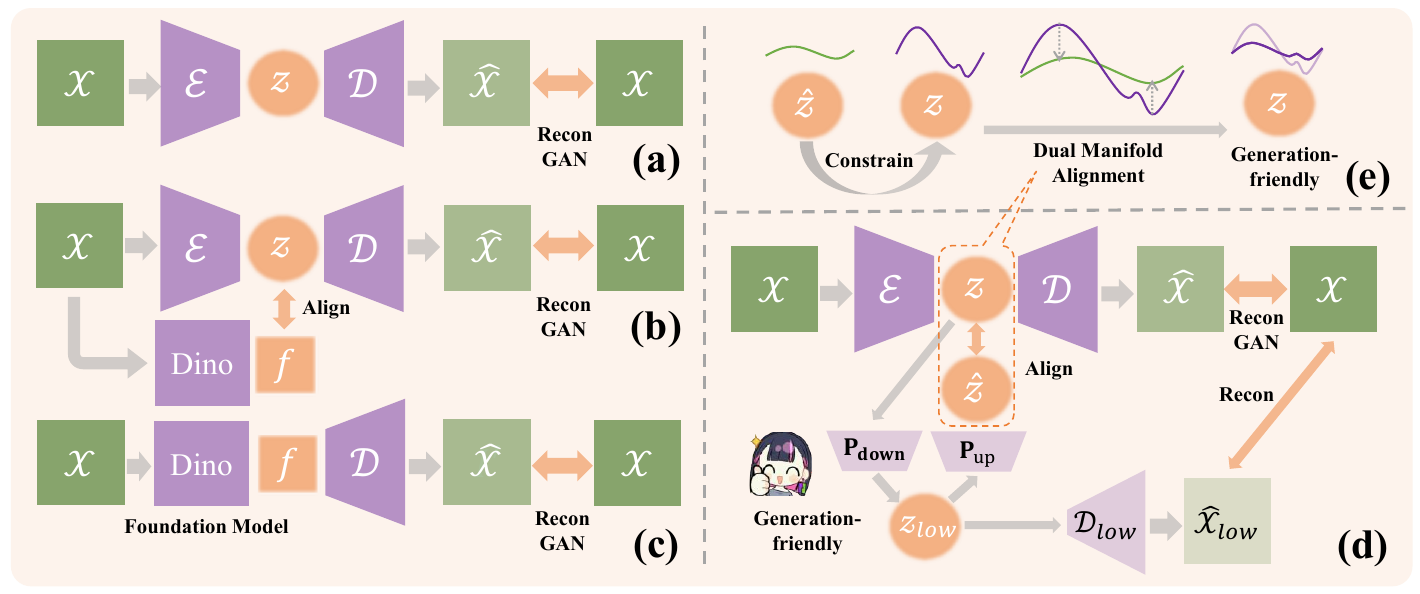}
    \vspace{-0.5em}
    \caption{\textbf{Comparison of reconstruction paradigms in LDMs.} 
Unlike \textbf{(a)} vanilla VAEs~\cite{rombach2022high} that suffer from unconstrained, rugged high-dimensional manifolds, or \textbf{(b)} foundation-model-aligned~\cite{yao2025reconstruction} and \textbf{(c)} foundation-model-based~\cite{zheng2026diffusion} VAEs that rely on external priors (\eg, Dino~\citep{oquab2023dinov2}) with limited generalization ability, \textbf{(d)} our foundation-model-free {\textbf{DM-VAE}} constructs an internal topological regularizer. Specifically, given an encoder $\mathcal{E}$ mapping the input to a high-dimensional latent $\bm{z}$ and a main decoder $\mathcal{D}$ for reconstruction, we compress $\bm{z}$ via $\mathcal{P}_{\text{down}}$ into a low-dimensional branch (comprising $\bm{z}_{\text{low}}$ and a lightweight decoder $\mathcal{D}_{\text{low}}$). By enforcing the alignment between $\bm{z}$ and the lifted $\hat{z}$, our method compels the high-dimensional manifold to inherit a smooth, generation-friendly geometry from $\bm{z}_{\text{low}}$. \textbf{(e)} Smooth \vs rugged manifolds, where ``smooth'' denotes a geometry dominated by low-frequency principal structures with bounded high-frequency oscillations along normal directions (\cf, Appendix~B).}
    \label{fig:method}
    \vspace{-1em}
\end{figure*}

Latent diffusion models (LDMs)~\cite{rombach2022high,labs2025flux,peebles2023scalable,ma2024sit} employ a continuous-valued variational autoencoder (VAE)~\cite{kingma2013auto,higgins2017beta} to project raw pixel images into a low-dimensional latent space. Operating within this compressed space, generative models (\eg, DiT~\cite{peebles2023scalable}) learn to iteratively generate valid in-distribution latents from random noise, which are subsequently decoded back into the image domain by the VAE.

Specifically, the LDM generative modeling pipeline consists of two stages: \textcolor{figpurple}{\textbf{1)} \textbf{Reconstruction}}: In the reconstruction stage (\cf, Figure~\ref{fig:method}(a)), the VAE first encodes images into latents (\ie, mapping image distribution to latent distribution) and then decodes these latents to reconstructed images. It usually jointly utilizes multiple training objectives to improve the visual quality of the reconstructed images, such as reconstruction~\cite{zhang2018unreasonable}, Kullback-Leibler (KL), and adversarial~\cite{goodfellow2020generative, esser2021taming} losses. \textcolor{figpurple}{\textbf{2)} \textbf{Generation}}: In the generation stage, generative models, \eg, flow-based models~\cite{lipman2022flow,liu2022flow,ma2024sit}, learn the velocity field to transport the prior distribution (\eg, a Gaussian distribution) towards the latent distribution. The training objective typically minimizes the regression errors between the predicted velocities and the ground truth velocities.
Crucially, since the generation stage operates within the latent space, {\textit{the overall generative performance is upper-bounded by the reconstruction quality.}}

Therefore, to push the upper bound of generative performance, enhancing reconstruction quality is essential. Due to the VAE's inherent information bottleneck~\cite{alemi2016deep},  a straightforward approach is expanding the dimension of the latent space (\eg, channels) to further minimize the reconstruction error. However, it inevitably introduces excessive degrees of freedom to the manifold corresponding to the latent distribution (\ie, latent manifold). Since there is no explicit structural constraint on the underlying latent manifold, this unregularized channel expansion often yields a rugged and ill-conditioned manifold~\cite{yao2025reconstruction}. Consequently, during generation, unconstrained high-frequency components disrupt velocity learning and easily cause generated latents to drift off the manifold, known as the {\textit{off-manifold}} problem~\cite{song2019generative}.

To address this off-manifold issue, recent approaches have increasingly relied on external foundation models (\eg, DINOv2~\cite{oquab2023dinov2}, MAE~\cite{he2022masked}). They either align latents with their features~\cite{yao2025reconstruction} (\cf, Figure~\ref{fig:method}(b)) or adopt these features directly as latents~\cite{zheng2026diffusion,singh2026improved} (\cf, Figure~\ref{fig:method}(c)). Specifically, their pre-trained features encode semantics that form a ``smooth'' reference manifold, acting as a \emph{topological regularizer}. These regularizers can effectively smooth the unconstrained latent manifold, easing subsequent generative training. However, tying the latent space to external priors inherently restricts the VAE's generalization capability~\citep{chefer2026self, bfl2025flux2rep}. Because frozen foundation models are bound by their pre-training distributions and semantic biases, they fail to capture faithful visual details on out-of-distribution (OOD) domains, resulting in severe reconstruction degradation (\cf, Figure~\ref{fig:1}).

\begin{figure}
    \centering
    \includegraphics[width=0.99\linewidth]{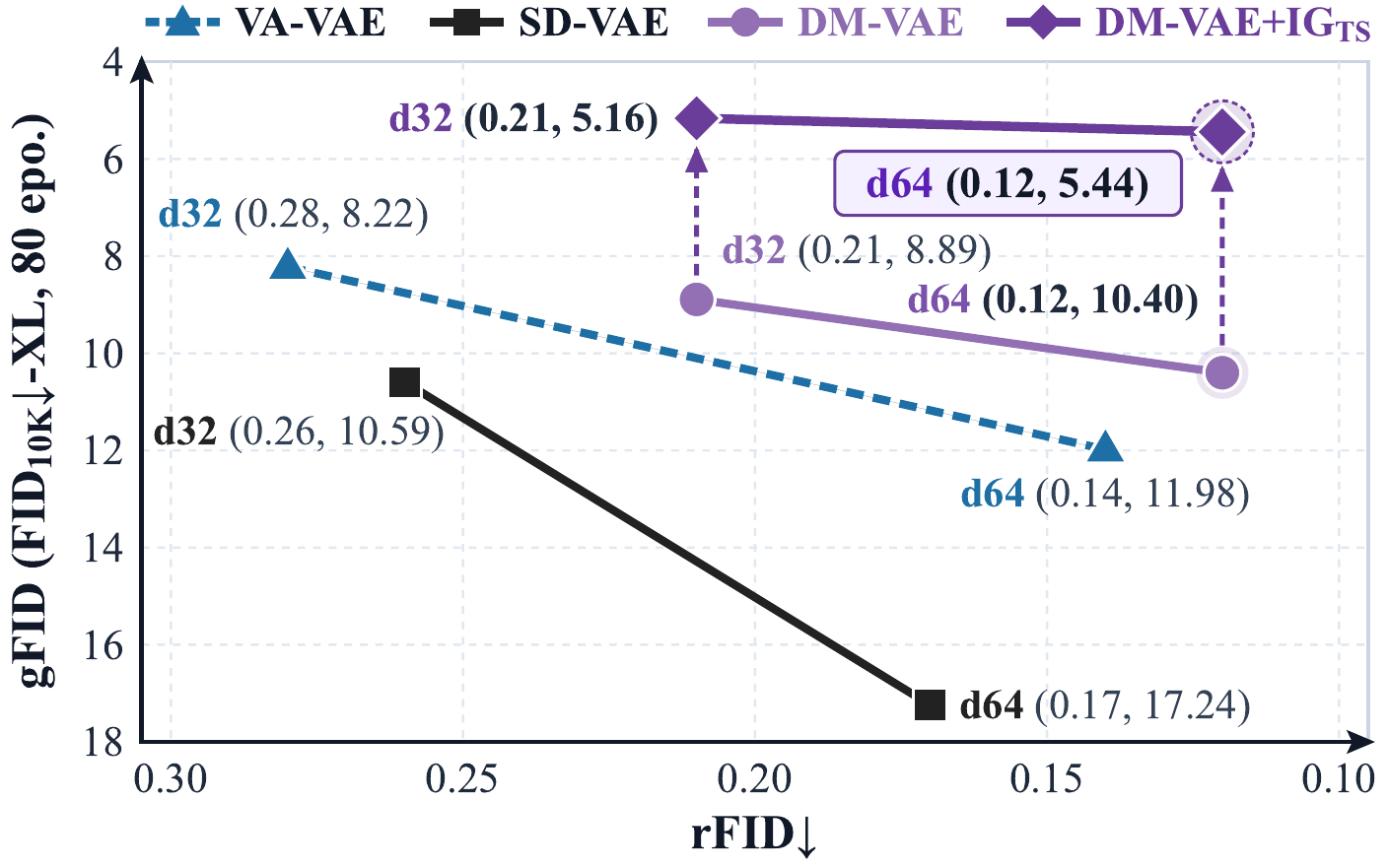}
    \vspace{-1em}
    \caption{\textbf{rFID vs. gFID$_{10K}$ comparison on DiT-XL (80 epochs).} When scaling from \texttt{d32} to \texttt{d64}, DM-VAE with $\text{IG}_{TS}$ effectively overcomes the reconstruction--generation trade-off \textbf{without external foundation models}.}
    \label{fig:3}
    \vspace{-1.5em}
\end{figure}

To this end, we propose a \textbf{\textcolor{figpurple}{dual-manifold VAE}} (\textbf{\textcolor{figpurple}{DM-VAE}}) framework to mitigate the off-manifold problem by \textit{leveraging internal representations as a topological regularizer}. Our design is motivated by a key observation: although low-dimensional latents sacrifice reconstruction fidelity, they consistently exhibit superior generative learnability~\citep{yao2025reconstruction} (\eg, SD-VAE~\citep{rombach2022high} in Figure~\ref{fig:3}), reflecting an inherently smoother underlying manifold~\citep{xu2026makingreconstructionfidpredictive}. Building on this insight, DM-VAE branches a shared encoder into two coupled latent spaces (\cf, Figure~\ref{fig:method}(d)): a high-dimensional space ($\bm{z}$) decoded by the main decoder for high-fidelity reconstruction, and a low-dimensional space ($\bm{z}_{\text{low}}$) (projected via $\mathcal{P}_{\text{down}}$ and supervised by a lightweight decoder $\mathcal{D}_{\text{low}}$) that captures a smooth topological ``skeleton''. To prevent unconstrained optimization on the high-dimensional latent manifold, we lift $\bm{z}_{\text{low}}$ back to the high-dimensional space via $\mathcal{P}_{\text{up}}$ to obtain $\hat{\bm{z}}$, and minimize its discrepancy with $\bm{z}$. This dual-manifold alignment compels the high-dimensional latent space to inherit the smooth geometry of its low-dimensional counterpart, smoothing the rugged manifold (\cf, Figure~\ref{fig:method}(e)).

Under our DM-VAE framework, the extracted topological skeleton naturally provides a structural prior for the generation stage.
To further facilitate generation within this regularized latent space, we introduce \textcolor{figpurple}{\textbf{topological representation alignment (TRA)}} as an additional supervision signal. Specifically, we align the intermediate hidden features of the generative model with the low-dimensional latents during training. Given that the low-dimensional manifold captures the principal structural components of its high-dimensional counterpart, this alignment serves as an explicit structural prior, guiding the model to learn a more accurate velocity field within the high-dimensional space. 

Beyond benefiting the training, these aligned intermediate features also enable an effective guidance strategy at the inference stage, \textcolor{figpurple}{\textbf{internal guidance with topological skeleton ($\text{IG}_{TS}$)}}. Although recent internal guidance (IG)~\cite{zhou2026guiding} provides an effective paradigm by using intermediate features, its potential is bottlenecked when directly applied to the entire latent space, as it tends to damage high-frequency details and exacerbate the off-manifold problem. In contrast, since our intermediate features are explicitly aligned with the low-dimensional latents, they naturally capture the clean topological skeleton. Building on this, $\text{IG}_{TS}$ separates the predicted high-dimensional latents into the low-dimensional topological skeleton and high-frequency details. We then use the aligned intermediate features to guide only the skeleton part, and add the unchanged high-frequency details back. In this way, $\text{IG}_{TS}$ improves structural quality while keeping the generated latents on the manifold, all within a single forward pass.

To validate the effectiveness of our method, we have conducted extensive experiments on ImageNet-1K~\citep{deng2009imagenet} at $256\times256$ resolution using DiT~\citep{peebles2023scalable}, following the LightningDiT~\citep{yao2025reconstruction} training settings. As shown in Figure~\ref{fig:3}, our approach combined with $\text{IG}_{TS}$ breaks the reconstruction--generation dilemma when scaling to higher-dimensional latent spaces (\texttt{d32} to \texttt{d64}), consistently outperforming other baselines.   
\textcolor{figpurple}{\textbf{Without relying on external foundation models}}, our DM-VAE (\texttt{f16d64}) achieves a reconstruction FID (rFID) of \textcolor{figpurple}{\textbf{0.12}} within 50 reconstructive training epochs and a 400-epoch generation $\text{FID}_{50K}$ of 1.28 with $\text{IG}_{TS}$ alone (1.20 under CFG). Our \texttt{f16d32} variant reaches \textcolor{figpurple}{\textbf{1.67}} under CFG at only 64 epochs and further advances to \textcolor{figpurple}{\textbf{1.25}} with $\text{IG}_{TS}$ alone (\textcolor{figpurple}{\textbf{1.17}} under CFG) at 400 epochs.

In summary, our main contributions are threefold: 
\textcolor{figpurple}{\textbf{1)}} We propose dual-manifold VAE (DM-VAE), a foundation-model-free VAE framework that regularizes rugged high-dimensional latent manifolds via a smooth low-dimensional topological skeleton.
\textcolor{figpurple}{\textbf{2)}} We introduce topological representation alignment (TRA) for training and internal guidance with topological skeleton ($\text{IG}_{TS}$) for inference, leveraging the skeleton for structural supervision and single-pass guidance while keeping high-frequency details intact. 
\textcolor{figpurple}{\textbf{3)}} Extensive experiments demonstrate superior reconstruction and generation performance in our DM-VAE framework.


\section{Related Work}

\noindent\textbf{Variational Autoencoder (VAE).} 
Continuous-valued VAEs~\citep{kingma2013auto,rezende2014stochastic,burda2015importance,doersch2016tutorial} serve as the cornerstone of latent diffusion models (LDMs)~\citep{labs2025flux,ma2024sit, rombach2022high,esser2021taming,peebles2023scalable,esser2024scaling,chen2026bi,li2026spar}. While vanilla VAEs were originally designed as end-to-end generative models that decode images directly from a latent prior distribution, they suffer from a structural mismatch between this predefined prior and the aggregated posterior distribution~\citep{hoffman2016elbo,tomczak2018vae}. Consequently, direct sampling from the vanilla VAE prior often yields blurry and low-fidelity images. To address this, LDMs repurpose the VAE as a reconstructive compressor, mapping high-dimensional pixel data into a compact latent space to facilitate subsequent generative training. Typically, training reconstructive VAE relies on reconstruction losses (\eg, $L_1$ and LPIPS~\citep{zhang2018unreasonable}) and adversarial~\citep{goodfellow2020generative,esser2021taming} loss, paired with a Kullback-Leibler (KL) divergence penalty to regularize the latent space. To preserve information for high reconstruction fidelity, the KL weight is usually set to a small value (\eg, $\beta = 10^{-6}$), exerting minimal influence on the latent distribution. However, without further structural regularization beyond this weak KL constraint, the learned latent space is prone to off-manifold issues during generation.

\noindent\textbf{Representation Alignment in LDMs.} Incorporating external or internal representation alignment objectives in latent generative modeling has proven effective in accelerating the training process. External alignment approaches, such as VA-VAE~\citep{yao2025reconstruction}, REPA~\citep{yu2024representation}, and REPA-E~\citep{leng2025repa}, leverage external foundation models (\ie, DINOv2~\citep{oquab2023dinov2}, MAE~\citep{he2022masked}, CLIP~\citep{radford2021learning}, SigLIP~\citep{tschannen2025siglip}) to guide the alignment of either the latent space or the intermediate features of DiTs. However, relying on external foundation models can lead to generalization issues. To address this, internal alignment methods (\eg, SRA~\citep{jiang2025no} and Self-Flow~\citep{chefer2026self}) instead align representations derived from different views of the same training sample in generative training. Unlike these existing approaches, our proposed dual-manifold VAE can optimize the latent space itself through internal alignment, thereby further facilitating generative training.

\section{Methodology}
\label{sec:method}

In this section, we first review the preliminary formulation of latent diffusion models (Sec.~\ref{sec:preliminaries}) and analyze the inherent off-manifold problem in unconstrained high-dimensional latent spaces. To mitigate this, we introduce the dual-manifold VAE framework that {{internalizes topological regularization without relying on external foundation models}}. We detail the {{dual-manifold alignment}} for the reconstruction stage (Sec.~\ref{sec:reconstruction}). Additionally, we introduce {topological representation alignment (TRA)} for the subsequent generation stage, along with an inference strategy, {internal guidance with topological skeleton (IG$_{TS}$)}  (Sec.~\ref{sec:generation}).

\subsection{Preliminaries: Latent Diffusion Models}
\label{sec:preliminaries}
Latent diffusion models (LDMs) decouple the image generation pipeline into two distinct stages: \textcolor{figpurple}{\textbf{1)}} In the reconstruction stage, given an image $\bm{x} \in \mathcal{X}$, a continuous-valued VAE utilizes its encoder $\mathcal{E}$ to map the image into a low-dimensional latent representation $\bm{z} = \mathcal{E}(\bm{x})$, and the decoder $\mathcal{D}$ to reconstruct the image $\hat{\bm{x}} = \mathcal{D}(\bm{z})$. The reconstructive VAE is optimized using a combination of reconstruction loss $\mathcal{L}_{\text{rec}}$ (\eg, $L_1$ and LPIPS~\citep{zhang2018unreasonable}), a Kullback-Leibler (KL) divergence penalty $\mathcal{L}_{\text{KL}}$ towards a standard Gaussian prior, and an adversarial loss $\mathcal{L}_{\text{adv}}$~\citep{goodfellow2020generative,esser2021taming}:
\begin{equation}
\label{eq:vae_base}
\mathcal{L}_{\text{VAE}}(\bm{x}, \hat{\bm{x}}) = \mathcal{L}_{\text{rec}}(\bm{x}, \hat{\bm{x}}) + \beta \mathcal{L}_{\text{KL}} + \lambda_{\text{adv}} \mathcal{L}_{\text{adv}}(\hat{\bm{x}}),
\end{equation}
where $\beta$ and $\lambda_{\text{adv}}$ denote the loss weights. \textcolor{figpurple}{\textbf{2) }} In the generation stage, flow matching models operate within the constructed latent space $\mathcal{Z}$. Specifically, let $\bm{z}_1 = \bm{z} \in \mathcal{Z}$ denote the encoded target latent and $\bm{z}_0 \sim \mathcal{N}(\bm{0}, \mathbf{I})$ denote standard Gaussian noise. Flow matching constructs a probability path via a linear interpolation: $\bm{z}_t = t \bm{z}_1 + (1 - t) \bm{z}_0$ for $t \in [0, 1]$. To transport the prior distribution towards the latent distribution, the generative model learns a neural velocity field $\bm{v}_\theta(\bm{z}_t, t)$ by regressing the constant ground-truth vector field $\bm{u}_t = \bm{z}_1 - \bm{z}_0$. The training objective minimizes:
\begin{equation}
\mathcal{L}_{\text{FM}} = \mathbb{E}_{t, \bm{z}_0, \bm{z}_1} \left[ \| \bm{v}_\theta(\bm{z}_t, t) - (\bm{z}_1 - \bm{z}_0) \|^2 \right],
\end{equation}
 where $t \sim \mathcal{U}(0,1)$. If the underlying latent manifold is ill-conditioned, unconstrained high-frequency components disrupt velocity regression and cause generated latents to drift off the manifold (\ie, the off-manifold problem), degrading generation in high-dimensional latent spaces.
 

\subsection{Dual-Manifold Variational Autoencoders}
\label{sec:reconstruction}
\noindent\textbf{Off-Manifold Analysis.}
\label{sec:3.2.1}
As discussed in Section~\ref{sec:intro}, the overall generative performance is inherently upper-bounded by reconstruction quality. To push this bound, increasing the latent channel dimension $C$ is a standard practice. From an information-theoretic perspective~\citep{cover1991elements,alemi2018fixing}, the minimum reconstruction distortion $\varepsilon_{\text{recon}}$ is limited by the latent representation capacity. Since the latent spatial resolution ($H \times W$) is typically fixed, this capacity is dominated by $C$, yielding an exponential theoretical bound\footnote{The detailed derivation is provided in Appendix~A.}: 
$\varepsilon_{\text{recon}} \ge \alpha \exp(-\gamma \cdot C),$ where $\alpha$ and $\gamma$ are positive constants determined by the intrinsic dimension and signal variance of the raw image manifold (manifold assumption~\citep{chapelle2009semi,li2026back}). This inequality explicitly demonstrates that expanding the channel dimension is the most direct approach to minimize reconstruction error. However, this unregularized expansion inevitably introduces excessive degrees of freedom to the latent manifold. Mapping the intrinsically $d$-dimensional image manifold $\mathcal{M}_X$ into a high-dimensional space $\mathbb{R}^{C \times H \times W }$ forms a latent manifold $\mathcal{M}_Z$ with an exploding codimension $c = C \times H \times W - d$. Without any constraint, the encoder tends to exploit these excessive degrees of freedom. It may scatter substantial high-frequency energy across the $c$ normal directions to minimize pixel and adversarial errors, yielding an ill-conditioned latent manifold. During flow matching, the high-variance high-frequency components severely interfere with low-frequency structural velocity learning (\cf, Appendix~B), causing the off-manifold problem.


\noindent\textbf{Dual-Manifold Alignment.} While VAEs with low-dimensional latents exhibit degraded reconstruction quality compared to their high-dimensional counterparts, they typically achieve better generation~\citep{yao2025reconstruction}. This indicates that {\textit{the low-dimensional latent manifold is inherently smoother and more generation-friendly}}~\citep{xu2026makingreconstructionfidpredictive}. Motivated by this, a straightforward idea is to leverage the low-dimensional latent manifold as a topological regularizer, compelling the high-dimensional manifold to inherit this property. 

To achieve this, we design a dual-manifold VAE framework that simultaneously constructs two latent spaces sharing the encoder $\mathcal{E}$. Given an input image $\bm{x} \in \mathcal{X}$, the encoder first maps it into a high-dimensional latent $\bm{z} = \mathcal{E}(\bm{x}) \in \mathbb{R}^{C \times H \times W}$, which is decoded by the main decoder $\mathcal{D}$ into a high-fidelity reconstruction $\hat{\bm{x}} = \mathcal{D}(\bm{z})$. In parallel, we compress $\bm{z}$ into a low-dimensional latent $\bm{z}_{\text{low}} \in \mathbb{R}^{C_{\text{low}} \times H \times W}$ ($C_{\text{low}} < C$) via a linear down-projection layer $\mathcal{P}_{\text{down}}$ (\ie, a $1 \times 1$ convolution), followed by a {\textit{lightweight decoder}} $\mathcal{D}_{\text{low}}$ to output a structural reconstruction:
\begin{equation}
    \bm{z}_{\text{low}} = \mathcal{P}_{\text{down}}(\bm{z}), \quad \hat{\bm{x}}_{\text{low}} = \mathcal{D}_{\text{low}}(\bm{z}_{\text{low}}).
\end{equation}
While the main branch ($\hat{\bm{x}}$) is optimized with the full reconstructive objective $\mathcal{L}_{\text{VAE}}(\bm{x}, \hat{\bm{x}})$, the low-dimensional branch ($\hat{\bm{x}}_{\text{low}}$) is supervised solely by reconstruction loss  $\mathcal{L}_{\text{rec}}(\bm{x}, \hat{\bm{x}}_{\text{low}})$. Constrained by the bottleneck $C_{\text{low}}$ and free from high-frequency adversarial training, $\bm{z}_{\text{low}}$ forms {\textit{a smooth manifold that captures the clean topological ``skeleton''}} from $\bm{z}$ (\ie, principal structural components). Rather than allowing the high-dimensional manifold to be optimized without any restriction, we treat $\bm{z}_{\text{low}}$ as an internal topological regularizer. Concretely, we project $\bm{z}_{\text{low}}$ back into the high-dimensional space via a corresponding up-projection layer $\mathcal{P}_{\text{up}}$ to obtain the projected topological skeleton $\hat{\bm{z}} = \mathcal{P}_{\text{up}}(\bm{z}_{\text{low}}) \in \mathbb{R}^{C \times H \times W}$, and minimize the $L_2$ distance between $\bm{z}$ and $\hat{\bm{z}}$ (\ie, {dual-manifold alignment}:
$
    \mathcal{L}_{\text{DM}} = \mathbb{E}_{\bm{x}} \left[ \| \bm{z} - \hat{\bm{z}} \|_2^2 \right])
$\footnote{In practice, $\mathcal{L}_{\text{DM}}$ is computed on the posterior mean via $\mathbb{E}_{\bm{x}} [ 0.5 \| \bm{z} - \text{sg}[\hat{\bm{z}}] \|_2^2 + 0.5 \| \text{sg}[\bm{z}] - \hat{\bm{z}} \|_2^2 ]$ ($\text{sg}[\cdot]$: stop-gradient), paired with a lightweight orthogonality penalty on $\mathcal{P}_{\text{down}}, \mathcal{P}_{\text{up}}$ (\cf, Appendix~C).}.
The total training objective is:
\begin{equation}
    \mathcal{L}_{\text{dual}} = \mathcal{L}_{\text{VAE}}(\bm{x}, \hat{\bm{x}}) + \lambda_{\text{low}} \mathcal{L}_{\text{rec}}(\bm{x}, \hat{\bm{x}}_{\text{low}}) + \lambda_{\text{DM}} \mathcal{L}_{\text{DM}},
\end{equation}
where $\lambda_{\text{low}}$ and $\lambda_{\text{DM}}$ are coefficients. Through $\mathcal{L}_{\text{DM}}$, DM-VAE suppresses transverse high-frequency residual energy and confines $\bm{z}$ within a \textit{narrow tubular neighborhood} around lifted $\bm{z}_{\text{low}}$, decoupling structural velocity learning from high-frequency interference and mitigating the off-manifold problem (\cf, Appendix~B).

\begin{figure}
    \centering
    \includegraphics[width=0.97\linewidth]{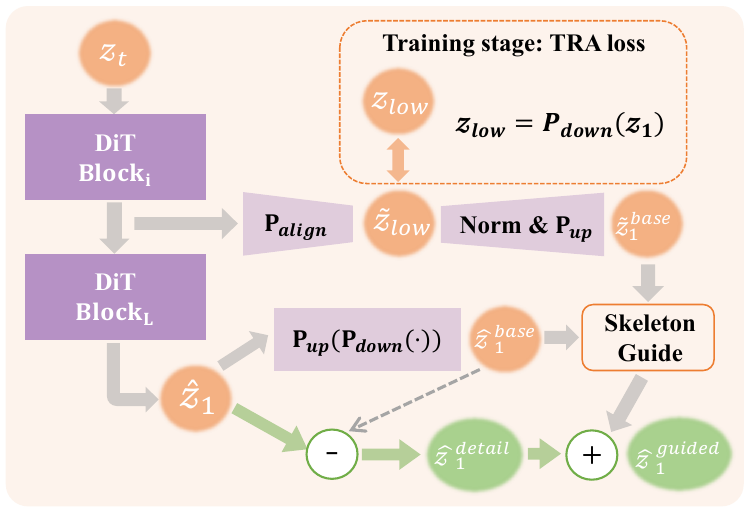}
    \vspace{-1em}
    \caption{\textbf{Overview of topological representation alignment (TRA) and internal guidance with topological skeleton ($\text{IG}_{TS}$).}}
\label{fig:ig_ts}
\end{figure}

\subsection{Topology-Guided Generation and Inference}
\label{sec:generation}

\noindent\textbf{Topological Representation Alignment.}
Building on the dual-manifold VAE (DM-VAE) framework, we naturally utilize the low-dimensional latent $\bm{z}_{\text{low}}$ to guide generative training in the main high-dimensional latent space (\cf, Figure~\ref{fig:ig_ts}). Specifically, given an interpolated latent $\bm{z}_t$ at timestep $t$, we extract the intermediate hidden features $\bm{h}_t$ from the generative model and project them into the low-dimensional space via a lightweight projection head $\mathcal{P}_{\text{align}}$ (\eg, an MLP) to obtain $\tilde{\bm{z}}_{\text{low}} = \mathcal{P}_{\text{align}}(\bm{h}_t) \in \mathbb{R}^{C_{\text{low}} \times H \times W}$. We then align $\tilde{\bm{z}}_{\text{low}}$ with the target topological skeleton $\bm{z}_{\text{low}} = \mathcal{P}_{\text{down}}(\bm{z}_1)$ by maximizing their cosine similarity:
\begin{equation}
\label{eq:TRA}
    \mathcal{L}_{\text{TRA}} = - \mathbb{E}_{t, \bm{z}_0, \bm{z}_1} \left[ \frac{\bm{z}_{\text{low}} \cdot \tilde{\bm{z}}_{\text{low}}}{\| \bm{z}_{\text{low}} \|_2 \| \tilde{\bm{z}}_{\text{low}} \|_2} \right].
\end{equation}
The overall generative training objective is formulated as:
$
\label{eq:gen_total}
    \mathcal{L}_{\text{gen}} = \mathcal{L}_{\text{FM}} + \lambda_{\text{TRA}} \mathcal{L}_{\text{TRA}},
$
where $\lambda_{\text{TRA}}$ is a coefficient. Since $\bm{z}_{\text{low}}$ captures the principal structural components of $\bm{z}_1$ on a smooth manifold, $\mathcal{L}_{\text{TRA}}$ compels the intermediate layers to establish a reliable structural representation early in the network. This explicit structural prior eases the subsequent regression of high-dimensional details, guiding the model to learn a more accurate velocity field $\bm{v}_\theta(\bm{z}_t, t)$.

\begin{algorithm}[t]
\caption{{\textcolor{figpurple}{Internal Guidance with Topological Skeleton}}}
\label{alg:1}
\small
\linespread{0.92}\selectfont
\begin{algorithmic}[1]
\REQUIRE Model $\bm{v}_\theta$ with $\mathcal{P}_{\text{align}}$, DM-VAE $\{\mathcal{P}_{\text{down}}, \mathcal{P}_{\text{up}}, \mathcal{D}\}$, scale $\omega \ge 1$, interval $[t_{\min}, t_{\max}]$, steps $N$. \textbf{Ensure:} Image $\hat{\bm{x}}$.

\STATE Initialize $\bm{z}_0 \sim \mathcal{N}(\bm{0}, \mathbf{I})$, $\Delta t \leftarrow 1 / N$
\FOR{$i = 0, \dots, N - 1$}
    \STATE $t \leftarrow i / N$; \quad $\omega_t \leftarrow \omega \text{ if } t \in [t_{\min}, t_{\max}] \text{ else } 1$
    \STATE \textcolor{figgreen}{\textit{\% 1. Single forward pass for velocity and skeleton}}
    \STATE $\bm{v}_\theta(\bm{z}_t, t), \bm{h}_t \leftarrow \text{Forward}(\bm{z}_t, t)$; \quad $\tilde{\bm{z}}_{\text{low}} \leftarrow \mathcal{P}_{\text{align}}(\bm{h}_t)$
    \STATE $\hat{\bm{z}}_1 \leftarrow \bm{z}_t + (1 - t)\bm{v}_\theta(\bm{z}_t, t)$
    
    \STATE \textcolor{figgreen}{\textit{\% 2. Separate skeleton and high-frequency details}}
    \STATE $\hat{\bm{z}}_{\text{low}} \leftarrow \mathcal{P}_{\text{down}}(\hat{\bm{z}}_1)$; \ $\hat{\bm{z}}_1^{\text{base}} \leftarrow \mathcal{P}_{\text{up}}(\hat{\bm{z}}_{\text{low}})$; \ $\hat{\bm{z}}_1^{\text{detail}} \leftarrow \hat{\bm{z}}_1 - \hat{\bm{z}}_1^{\text{base}}$
    
    \STATE \textcolor{figgreen}{\textit{\% 3. Calibrate intermediate norm and guide the skeleton}}
    \STATE $\tilde{\bm{z}}_1^{\text{base}} \leftarrow \mathcal{P}_{\text{up}}\big( \| \hat{\bm{z}}_{\text{low}} \|_2 \cdot \tilde{\bm{z}}_{\text{low}} / \| \tilde{\bm{z}}_{\text{low}} \|_2 \big)$
    \STATE $\hat{\bm{z}}_1^{\text{guided}} \leftarrow \hat{\bm{z}}_1^{\text{base}} + (\omega_t - 1)\big(\hat{\bm{z}}_1^{\text{base}} - \tilde{\bm{z}}_1^{\text{base}}\big) + \hat{\bm{z}}_1^{\text{detail}}$, Eq.~\eqref{eq:ig_ts}
    
    \STATE \textcolor{figgreen}{\textit{\% 4. Update guided velocity and latent state}}
    \STATE $\bm{v}_{\text{guided}} \leftarrow (\hat{\bm{z}}_1^{\text{guided}} - \bm{z}_t) / (1 - t)$; \quad $\bm{z}_{t + \Delta t} \leftarrow \bm{z}_t + \bm{v}_{\text{guided}} \cdot \Delta t$
\ENDFOR
\RETURN $\hat{\bm{x}} = \mathcal{D}(\bm{z}_1)$
\end{algorithmic}
\end{algorithm}

\noindent\textbf{Internal Guidance with Topological Skeleton.}
Beyond benefiting generative training, the aligned intermediate prediction $\tilde{\bm{z}}_{\text{low}}$ also enables an effective inference strategy, internal guidance with topological skeleton (IG$_{TS}$). As shown in Figure~\ref{fig:ig_ts}, at timestep $t$, the model outputs both the predicted velocity $\bm{v}_\theta(\bm{z}_t, t)$ and $\tilde{\bm{z}}_{\text{low}}$ in a single forward pass, yielding the one-step target estimate $\hat{\bm{z}}_1 = \bm{z}_t + (1 - t)\bm{v}_\theta(\bm{z}_t, t)$. While standard internal guidance (IG)~\citep{zhou2026guiding} extrapolates between final and intermediate predictions across the entire latent space and may corrupt high-frequency details, IG$_{TS}$ uses our dual-manifold VAE projections $\mathcal{P}_{\text{down}}$ and $\mathcal{P}_{\text{up}}$ to separate $\hat{\bm{z}}_1$ into the topological skeleton $\hat{\bm{z}}_1^{\text{base}}$ and high-frequency details $\hat{\bm{z}}_1^{\text{detail}}$:
$
\label{eq:decompose}
    \hat{\bm{z}}_{\text{low}} = \mathcal{P}_{\text{down}}(\hat{\bm{z}}_1), \quad \hat{\bm{z}}_1^{\text{base}} = \mathcal{P}_{\text{up}}(\hat{\bm{z}}_{\text{low}}), \quad \hat{\bm{z}}_1^{\text{detail}} = \hat{\bm{z}}_1 - \hat{\bm{z}}_1^{\text{base}}.
$
Since $\tilde{\bm{z}}_{\text{low}}$ is trained only with cosine similarity, we calibrate its norm using $\hat{\bm{z}}_{\text{low}}$ and project it to obtain the intermediate skeleton $\tilde{\bm{z}}_1^{\text{base}}$:
$
\label{eq:inter_skeleton}
    \tilde{\bm{z}}_1^{\text{base}} = \mathcal{P}_{\text{up}}\left( \| \hat{\bm{z}}_{\text{low}} \|_2 \cdot {\tilde{\bm{z}}_{\text{low}}}/{\| \tilde{\bm{z}}_{\text{low}} \|_2} \right).
$
We then apply skeleton guidance (\ie, $\hat{\bm{z}}_1^{\text{base}} + (\omega - 1)\left(\hat{\bm{z}}_1^{\text{base}} - \tilde{\bm{z}}_1^{\text{base}}\right)$) and add the high-frequency details back:
\begin{equation}
\label{eq:ig_ts}
    \hat{\bm{z}}_1^{\text{guided}} = \hat{\bm{z}}_1^{\text{base}} + (\omega - 1)\left(\hat{\bm{z}}_1^{\text{base}} - \tilde{\bm{z}}_1^{\text{base}}\right) + \hat{\bm{z}}_1^{\text{detail}},
\end{equation}
where $\omega \ge 1$ is the internal guidance scale. The guided velocity is then updated as $\bm{v}_{\text{guided}}(\bm{z}_t, t) = (\hat{\bm{z}}_1^{\text{guided}} - \bm{z}_t) / (1 - t)$. 
In this way, IG$_{TS}$ improves structural quality while mitigating off-manifold extrapolation drift by leaving the high-frequency details unperturbed. The detailed IG$_{TS}$ process with time interval~\citep{kynkaanniemi2024applying} is shown in Algorithm~\ref{alg:1}.

\begin{figure*}
    \centering
    \includegraphics[width=0.99\linewidth]{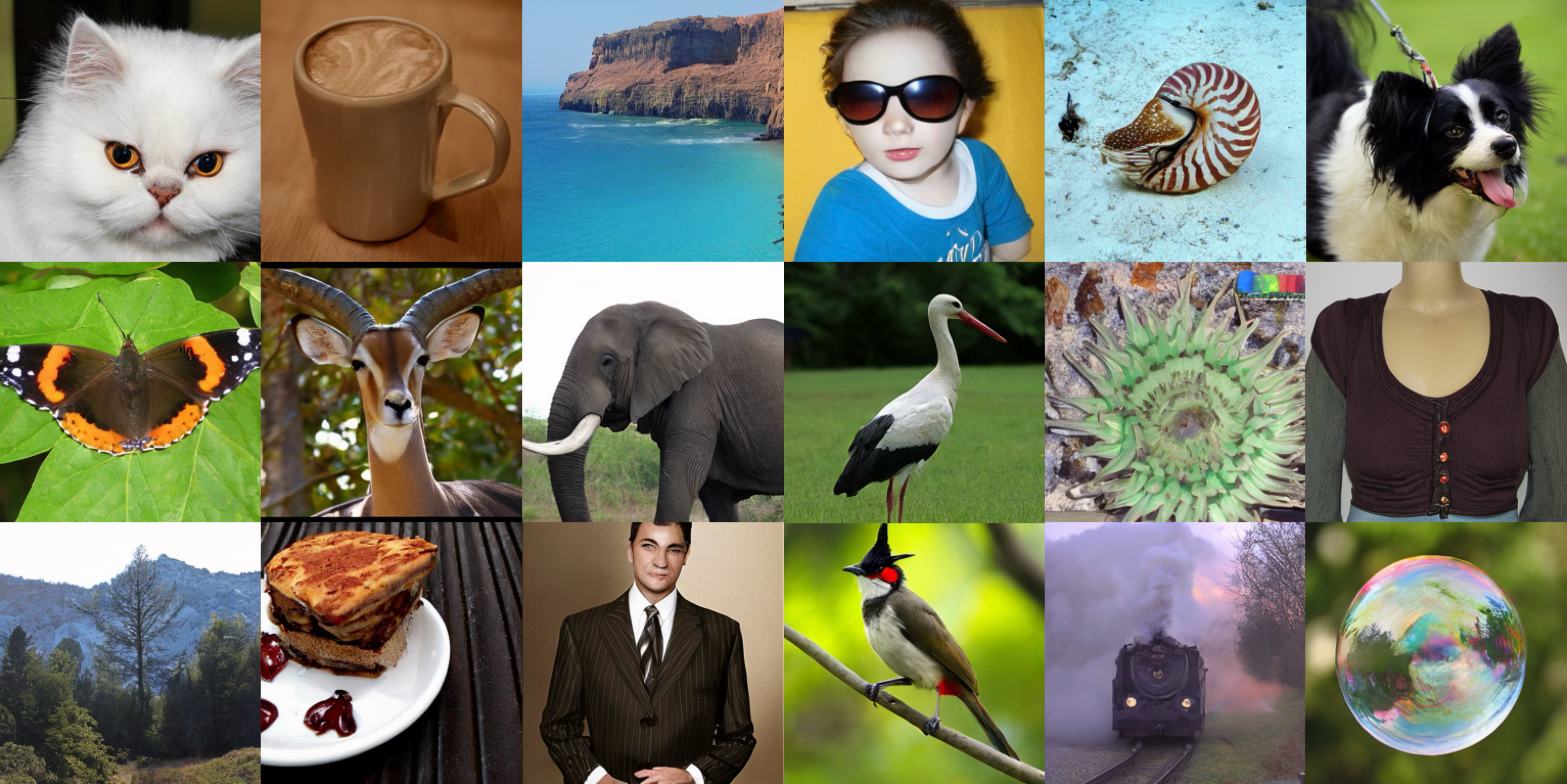}
    \vspace{-0.5em}
    \caption{\textbf{Qualitative generation results} on ImageNet using DiT-XL with our foundation-model-free DM-VAE (\texttt{f16d32}) and $\text{IG}_{TS}$ at only \textbf{64 training epochs} (\textbf{without CFG}).}
    \label{fig:samples}
    \vspace{-1em}
\end{figure*}

\section{Experiments}

\subsection{Experimental Settings}


\noindent\textbf{Datasets, Baselines, and Evaluation Metrics.} 
We evaluate both image reconstruction and class-conditional generation on ImageNet-1K~\citep{deng2009imagenet} at a resolution of $256 \times 256$. 
Following the categorization in our main comparisons, we benchmark against two categories of baselines: \textcolor{figpurple}{\textbf{1)}} {\textit{Vanilla latent diffusion models}} (without external foundation models), including DiT~\citep{peebles2023scalable}, MaskDiT~\citep{zheng2024fasttrainingdiffusionmodels}, SiT~\citep{ma2024sit}, TREAD~\citep{krause2025tread}, and MDTv2~\citep{gao2023mdtv2}; 
\textcolor{figpurple}{\textbf{2)}} {\textit{Foundation-model-aligned and guided LDMs}}, which regularize either the VAE latent space (\eg, MAE-VAE and VA-VAE with LightningDiT~\citep{yao2025reconstruction}, RAE~\citep{zheng2026diffusion}, IG (VA-VAE)~\citep{zhou2026guiding}) or the intermediate generative features (\eg, REPA~\citep{yu2024representation}, REPA-E~\citep{leng2025repa}, and REG~\citep{wu2026representation}). 
To verify versatility across generative architectures, we integrate our DM-VAE and IG$_{TS}$ into DiT~\citep{peebles2023scalable}, following the LightningDiT~\citep{yao2025reconstruction} training settings across both Base (B) and Extra-Large (XL) scales. 
For evaluation metrics, we adopt reconstruction FID (rFID), PSNR, LPIPS~\citep{zhang2018unreasonable}, and SSIM~\citep{wang2004image} on both the ImageNet-1K $50\text{K}$ validation set (in-domain) and the Anime-Faces-256 benchmark~\citep{danbooru2021} ($5\text{K}$, OOD) for the first stage, and reported Fr\'echet Inception Distance (FID)~\citep{heusel2017gans} ($\text{FID}_{10\text{K}}$ and $\text{FID}_{50\text{K}}$), spatial FID (sFID)~\citep{nash2021generating}, Inception Score (IS), and Precision-Recall~\citep{kynkaanniemi2019improved} for generation quality.

\noindent\textbf{Implementation Details.} 
\textcolor{figpurple}{\textbf{1)}} Reconstruction stage: During reconstructive training, we employ adaptive weighting coefficients to dynamically balance the loss terms. For our foundation-model-free variant (\textcolor{figpurple}{\textbf{DM-VAE}}), we train two configurations within 50 epochs with a downsampling factor of $16$: \texttt{f16d32} ($C=32$, low-dimensional branch $C_{\text{low}}=16$) and \texttt{f16d64} ($C=64$, low-dimensional branch $C_{\text{low}}=32$). To demonstrate orthogonality with external representations, we also train a DINOv2-integrated~\citep{oquab2023dinov2} variant (\textcolor{figpurple}{\textbf{DM-VAE$_{\textit{Dino}}$}}) under the \texttt{f16d32} setting\footnote{Specifically, we incorporate DINOv2 following the visual foundation (VF) alignment loss formulation in VA-VAE~\citep{yao2025reconstruction}.}. All training hyperparameters follow VA-VAE~\citep{yao2025reconstruction} (\eg, a global batch size of $256$ and a learning rate of $1\times 10^{-4}$). Detailed training settings are provided in Appendix~C.
\textcolor{figpurple}{\textbf{2)}} Generation stage: We conduct generative training on ImageNet-1K at a resolution of $256 \times 256$ using DiT\footnote{In this paper, unless otherwise specified, we adopt the modernized LightningDiT~\citep{yao2025reconstruction} architecture and refer to it as DiT~\citep{peebles2023scalable} for brevity.}, adhering to the official default training protocols~\citep{yao2025reconstruction,zhou2026guiding} (\eg, a global batch size of $1024$ and a learning rate of $2\times 10^{-4}$). We then employ a 250-step Euler solver for sampling. 

\subsection{Main Results}
\label{sec:main_results}

\noindent\textbf{In-Domain and OOD Reconstruction.} 
Table~\ref{tab:recon_ood} compares reconstruction quality on ImageNet-1K (in-domain) and Anime-Faces-256~\citep{danbooru2021} (zero-shot OOD). 
\textcolor{figpurple}{\textbf{1)}} {\textit{In-domain fidelity:}} Decoding from frozen semantic encoders in RAE~\citep{zheng2026diffusion} and RAEv2~\citep{singh2026improved} severely degrades pixel fidelity ($19.20$ and $23.29\text{ dB}$ PSNR), while external alignment in MAE-VAE and VA-VAE~\citep{yao2025reconstruction} ($0.28$ rFID) slightly trails vanilla SD-VAE ($0.26$ rFID). Conversely, our {DM-VAE} preserves local details and improves rFID to \textcolor{figpurple}{\textbf{0.21}} (\texttt{f16d32}) and \textcolor{figpurple}{\textbf{0.12}} (\texttt{f16d64}, \textcolor{figpurple}{\textbf{30.76}}$\text{ dB}$ PSNR). 
\textcolor{figpurple}{\textbf{2)}} {\textit{OOD generalization:}} On OOD anime illustrations, foundation-model-based tokenizers suffer from domain shift, with RAE/RAEv2 collapsing to $9.85$/$5.09$ rFID and VA-VAE/MAE-VAE lagging behind SD-VAE in pixel and perceptual fidelity ($29.78$/$29.33\text{ dB}$ vs.\ $30.19\text{ dB}$ PSNR). Free from external pretraining biases, {DM-VAE} generalizes robustly, achieving \textcolor{figpurple}{\textbf{1.93}} rFID ($30.30\text{ dB}$ PSNR) at \texttt{f16d32} and \textcolor{figpurple}{\textbf{1.12}} rFID (\textcolor{figpurple}{\textbf{32.87}}$\text{ dB}$ PSNR, \textcolor{figpurple}{\textbf{0.95}} SSIM) at \texttt{f16d64}.

\begin{table}[t]
\centering
\setlength{\tabcolsep}{5.5pt}
\scriptsize
\resizebox{\columnwidth}{!}{%
\begin{tabular}{l | cccc}
\toprule
\textbf{Methods} & \textbf{rFID}$\downarrow$ & \textbf{PSNR}$\uparrow$ & \textbf{LPIPS}$\downarrow$ & \textbf{SSIM}$\uparrow$ \\
\midrule
\multicolumn{5}{l}{\textcolor{figgreen}{\textit{\textbf{In-domain reconstruction performance}}}} \\
\midrule
MAE-VAE (\texttt{f16d32})~\citep{yao2025reconstruction} & 0.28 & 28.33 & 0.091 & 0.80 \\
VA-VAE (\texttt{f16d32})~\citep{yao2025reconstruction}  & 0.28 & 27.96 & 0.096 & 0.79 \\
RAE~\citep{zheng2026diffusion}                          & 0.57   & 19.20    & 0.254    & 0.44   \\
RAEv2~\citep{singh2026improved}                         & 0.31   & 23.29    & 0.148    & 0.61   \\
\midrule
SD-VAE (\texttt{f16d32})~\citep{rombach2022high}        & 0.26 & 28.59 & 0.089 & 0.80 \\
\textcolor{figpurple}{DM-VAE (\texttt{f16d32})}         & 0.21 & 28.50   & 0.088    & 0.80   \\
\textcolor{figpurple}{DM-VAE (\texttt{f16d64})}         & \textbf{0.12} & \textbf{30.76} & \textbf{0.057} & \textbf{0.87} \\
\midrule
\multicolumn{5}{l}{\textcolor{figgreen}{\textit{\textbf{Out-of-distribution reconstruction performance}}}} \\
\midrule
MAE-VAE (\texttt{f16d32})~\citep{yao2025reconstruction} & 2.38 & 29.78   & 0.048 & 0.91 \\
VA-VAE (\texttt{f16d32})~\citep{yao2025reconstruction}  & 2.05 & 29.33    & 0.052 & 0.90 \\
RAE~\citep{zheng2026diffusion}                          & 9.85 & 17.53 & 0.239 & 0.50 \\
RAEv2~\citep{singh2026improved}                         & 5.09 & 22.20   & 0.120 & 0.70 \\
\midrule
SD-VAE (\texttt{f16d32})~\citep{rombach2022high}        & 2.08 & 30.19    & 0.046 & 0.92 \\
\textcolor{figpurple}{DM-VAE (\texttt{f16d32})}         & 1.93 & 30.30    & 0.046 & 0.92 \\
\textcolor{figpurple}{DM-VAE (\texttt{f16d64})}         & \textbf{1.12} & \textbf{32.87} & \textbf{0.027} & \textbf{0.95} \\
\bottomrule
\end{tabular}%
}
\vspace{-0.5em}
\caption{\textbf{In-domain (ImageNet-1K) and out-of-distribution (OOD) reconstruction performance comparison.} Without relying on external foundation models, our DM-VAE achieves state-of-the-art reconstruction fidelity across both benchmarks.}
\label{tab:recon_ood}
\vspace{-0.5em}
\end{table}

\noindent\textbf{Integration with DiT.} 
Table~\ref{tab:lightning} evaluates reconstruction and generation performance on DiT-B and DiT-XL (80/160 epochs, without CFG). 
\textcolor{figpurple}{\textbf{1)}} {\textit{Reconciling the reconstruction--generation dilemma without external priors:}} At \texttt{f16d32}, {DM-VAE} improves rFID over SD-VAE ($0.26 \to \textcolor{figpurple}{\textbf{0.21}}$) while lowering $\text{FID}_{10K}$ to $21.41$/$17.66$ (B) and $8.89$/$7.97$ (XL); pairing with $\text{IG}_{TS}$ further reaches \textcolor{figpurple}{\textbf{12.60}/\textbf{9.22}} (B) and \textcolor{figpurple}{\textbf{5.16}/\textbf{4.20}} (XL), surpassing DINOv2-aligned VA-VAE~\citep{yao2025reconstruction} ($20.34$/$15.73$ and $8.22$/$6.75$). More importantly, when expanding to \texttt{f16d64}, vanilla SD-VAE suffers severe off-manifold degradation ($\text{FID}_{10K}$ worsens to $17.24$/$12.38$ on XL despite $0.17$ rFID). Conversely, {DM-VAE} regularizes the 64-channel space to achieve \textcolor{figpurple}{\textbf{0.12}} rFID alongside $10.40$/$8.59$ $\text{FID}_{10K}$ on DiT-XL, and \textcolor{figpurple}{\textbf{5.44}/\textbf{4.26}} with $\text{IG}_{TS}$. Figure~\ref{fig:samples} confirms sharp textures and coherent structures at only 64 epochs without CFG. 
\textcolor{figpurple}{\textbf{2)}} {\textit{Synergy with external foundation models:}} Orthogonally combining DINOv2 in {DM-VAE}$_{\textit{Dino}}$ (\texttt{f16d32}) yields $0.27$ rFID and further accelerates convergence, reaching $6.99$/$5.97$ $\text{FID}_{10K}$ on DiT-XL and \textcolor{figpurple}{\textbf{4.86}/\textbf{4.33}} with $\text{IG}_{TS}$ (\textcolor{figpurple}{\textbf{6.72}/\textbf{6.51}} on DiT-B).

\noindent\textbf{Comparison with State-of-the-Art Methods.} 
In Table~\ref{tab:sota}, we benchmark our full framework against leading vanilla LDMs and foundation-model-guided LDMs on ImageNet-1K ($256\times256$). 
\textcolor{figpurple}{\textbf{1)}} {\textit{Foundation-model-free regime:}} Compared with vanilla LDMs such as vanilla DiT~\citep{peebles2023scalable}, SiT~\citep{ma2024sit}, and MDTv2~\citep{gao2023mdtv2} that require $1080\sim1600$ epochs to converge, DiT paired with our {DM-VAE} (\texttt{f16d32}) reaches an $\text{FID}_{50K}$ of $2.62$ ($\text{IG}_{TS}$ alone, w/o CFG) and $1.67$ (with CFG) at only 64 epochs, and achieves \textcolor{figpurple}{\textbf{1.25}} without CFG and \textcolor{figpurple}{\textbf{1.17}} with CFG at 400 epochs. Notably, when scaling to a 64-channel latent space (\texttt{f16d64}, $0.12$ rFID), our dual-manifold regularization and $\text{IG}_{TS}$ effectively prevent high-dimensional off-manifold degradation, achieving an $\text{FID}_{50K}$ of $3.01$/$2.10$ at 64 epochs, $1.71$/$1.43$ at 160 epochs, and \textcolor{figpurple}{\textbf{1.28}} (w/o CFG) / \textcolor{figpurple}{\textbf{1.20}} (w/ CFG) at 400 epochs. Consequently, both our \texttt{f16d32} and \texttt{f16d64} variants not only outperform all vanilla LDMs, but also surpass 800-epoch foundation-model-aligned methods including REPA~\citep{yu2024representation} ($1.42$), REPA-E~\citep{leng2025repa} ($1.26$), VA-VAE~\citep{yao2025reconstruction} ($1.35$), and RAE~\citep{zheng2026diffusion} ($1.28$).
\textcolor{figpurple}{\textbf{2)}} {\textit{Foundation-model-assisted regime:}} Furthermore, our framework is orthogonal to external priors; when combined with DINOv2 ({DM-VAE}$_{\textit{Dino}}$), it reaches an $\text{FID}_{50K}$ of \textcolor{figpurple}{\textbf{2.26}} (w/o CFG) within 60 epochs and further pushes the state-of-the-art $\text{FID}_{50K}$ (400 epochs) to \textcolor{figpurple}{\textbf{1.29}} without CFG and \textcolor{figpurple}{\textbf{1.11}} with CFG, outperforming IG~\citep{zhou2026guiding} (VA-VAE) ($1.19$ at 680 epochs).

\begin{table}[t]
\centering
\setlength{\tabcolsep}{3pt}
\scriptsize
\newcommand{\imp}[1]{$_{\textcolor{orange}{\textbf{#1}}}$}
\resizebox{\columnwidth}{!}{%
\begin{tabular}{lc | cc | cc}
\toprule
\multirow{2}{*}{\textbf{Methods}} & \multirow{2}{*}{\textbf{rFID}$\downarrow$} & \multicolumn{2}{c|}{\textbf{FID$_{10K}$$\downarrow$-B}} & \multicolumn{2}{c}{\textbf{FID$_{10K}$$\downarrow$-XL}} \\
& & 80 epo. & 160 epo. & 80 epo. & 160 epo. \\
\midrule
\multicolumn{6}{l}{\textcolor{figgreen}{\textit{\textbf{With foundation-model alignment}}}} \\
\midrule
MAE-VAE (\texttt{f16d32})~\citep{yao2025reconstruction} & 
0.28 & 
25.37 & 19.89 & 9.93 & 8.18 \\

VA-VAE (\texttt{f16d32})~\citep{yao2025reconstruction} & 
0.28 & 
20.34 & 15.73 & 8.22 & 6.75 \\

\textcolor{figpurple}{{DM-VAE}$_\textit{Dino}$ (\texttt{f16d32})} & 
\multirow{2}{*}{\textbf{0.27}} & 
14.06 & 12.11 & 6.99 & 5.97 \\

+ \textcolor{figpurple}{IG$_{TS}$} & 
& 
\textbf{6.72} & \textbf{6.51} & \textbf{4.86} & \textbf{4.33} \\

\midrule
\multicolumn{6}{l}{\textcolor{figgreen}{\textit{\textbf{Without foundation-model alignment}}}} \\
\midrule
SD-VAE (\texttt{f16d32})~\citep{rombach2022high} & 
0.26 & 
28.12 & 22.43 & 10.59 & 8.52 \\

\textcolor{figpurple}{DM-VAE (\texttt{f16d32})} & 
\multirow{2}{*}{{\textbf{0.21}}} & 
21.41\imp{-6.7} & 17.66\imp{-4.8} & 8.89\imp{-1.7} & 7.97\imp{-0.6} \\
+ \textcolor{figpurple}{IG$_{TS}$} & 
& 
\textbf{12.60}\imp{-15.5} & \textbf{9.22}\imp{-13.2} & \textbf{5.16}\imp{-5.4} & \textbf{4.20}\imp{-4.3} \\
\midrule

SD-VAE (\texttt{f16d64})~\citep{rombach2022high} & 
0.17 & 45.77 & 36.83 & 17.24 & 12.38 \\

\textcolor{figpurple}{DM-VAE (\texttt{f16d64})} &
\multirow{2}{*}{{\textbf{0.12}}} & 
31.39\imp{-14.4} & 24.96\imp{-11.9} & 10.40\imp{-6.8} & 8.59\imp{-3.8} \\
+ \textcolor{figpurple}{IG$_{TS}$} & 
& 
\textbf{20.14}\imp{-25.6} & \textbf{14.12}\imp{-22.7} & \textbf{5.44}\imp{-11.8} & \textbf{4.26}\imp{-8.1} \\
\bottomrule
\end{tabular}%
}
\vspace{-0.5em}
\caption{\textbf{Reconstruction and generation performance comparison on ImageNet} using \textbf{DiT}~\citep{peebles2023scalable} without CFG. Subscripts highlight $\text{FID}_{10K}$ improvements over the corresponding SD-VAE. Results are reported at 80 and 160 generative training epochs, with VAEs trained within 50 epochs. For IG$_{TS}$, we consistently apply a guidance scale of 1.7 across the timestep interval of $[0, 0.6]$.}
\vspace{-0.5em}
\label{tab:lightning}
\end{table}

\begin{table*}[t]
\scriptsize
\centering
\label{tab:diffusion_comparison}
{
\begin{tabular}{l c c ccccc ccccc}
\toprule
\multirow{2}{*}{\textbf{Method}} & \multirow{2}{*}{\textbf{Epochs}} & \multirow{2}{*}{\textbf{\#Params}} & \multicolumn{5}{c}{\textbf{Generation@256 w/o CFG}} & \multicolumn{5}{c}{\textbf{Generation@256 w/ CFG or Autoguidance}} \\
\cmidrule(lr){4-8} \cmidrule(lr){9-13}
& & & \textbf{FID}$\downarrow$ & \textbf{sFID}$\downarrow$ & \textbf{IS}$\uparrow$ & \textbf{Prec.}$\uparrow$ & \textbf{Rec.}$\uparrow$ & \textbf{FID}$\downarrow$ & \textbf{sFID}$\downarrow$ & \textbf{IS}$\uparrow$ & \textbf{Prec.}$\uparrow$ & \textbf{Rec.}$\uparrow$ \\
\midrule
\multicolumn{13}{l}{\textcolor{figgreen}{\textit{\textbf{Latent Diffusion with Self-supervised Representation Models}}}} \\
\midrule
\multirow{2}{*}{REPA~\citep{yu2024representation}} & 80 & \multirow{2}{*}{675M} & 7.90 & 5.06 & 122.6 & 0.70 & 0.65 & - & - & - & - & - \\
& 800 & & 5.90 & 5.69 & 157.8 & 0.70 & \textbf{0.69} & 1.42 & 4.70 & 305.7 & 0.80 & 0.65 \\
\midrule
\multirow{2}{*}{REPA-E~\citep{leng2025repa}} & 80 & \multirow{2}{*}{675M} & 3.46 & 4.17 & 159.8 & 0.77 & 0.63 & 1.67 & 4.12 & 266.3 & 0.80 & 0.63 \\
& 800 & & 1.83 & - & 217.3 & - & - & 1.26 & 4.11 & 314.9 & 0.79 & 0.66 \\
\midrule
\multirow{2}{*}{REG~\citep{wu2026representation}} & 80 & \multirow{2}{*}{677M} & 3.40 & - & 184.1 & - & - & 1.86 & 4.49 & \textbf{321.4} & 0.76 & 0.63 \\
& 480 & & 2.20 & 4.67 & 219.1 & 0.77 & 0.66 & 1.40 & 4.24 & 296.9 & 0.77 & 0.66 \\
\midrule
\multirow{2}{*}{VA-VAE~\citep{yao2025reconstruction}} & 64 & \multirow{2}{*}{675M} & 5.14 & 4.22 & 130.2 & 0.76 & 0.62 & 2.11 & 4.16 & 252.3 & \textbf{0.81} & 0.58 \\
& 800 & & 2.17 & 4.36 & 205.6 & 0.77 & 0.65 & 1.35 & 4.15 & 295.3 & 0.79 & 0.65 \\
\midrule
\multirow{3}{*}{DDT~\citep{wang2026ddt}} & 80 & \multirow{3}{*}{675M} & 6.62 & - & 135.2 & 0.69 & 0.67 & 1.52 & - & 263.7 & 0.78 & 0.63 \\
& 256 & & 6.30 & - & 146.7 & 0.68 & 0.68 & 1.31 & - & 308.1 & 0.78 & 0.62 \\
& 400 & & 6.27 & - & 154.7 & 0.68 & \textbf{0.69} & 1.26 & - & 310.6 & 0.79 & 0.65 \\
\midrule
\multirow{2}{*}{IG~\citep{zhou2026guiding} (VA-VAE)} & 60 & \multirow{2}{*}{675M} & 2.42 & \textbf{3.81} & 173.7 & \textbf{0.79} & 0.62 & - & - & - & - & - \\
& 680 & & 1.34 & 3.94 & 229.3 & 0.78 &  0.65 & 1.19 & 4.11 & 269.0 & 0.79 & 0.66 \\
\midrule
RAE (DiT$^{\text{DH}}$)~\citep{zheng2026diffusion} & 800 & 839M & 1.60 & 5.40 & {242.7} & \textbf{0.79} & 0.65 & 1.28 & 4.72 & 262.9 & 0.78 & \textbf{0.67} \\
\midrule
\multirow{3}{*}{\textcolor{figpurple}{\textbf{DiT-XL (DM-VAE$_{\textit{Dino}}$)}}} & 60 & \multirow{3}{*}{675M} & 2.26 & 4.81 & 200.3 & \textbf{0.79} & 0.60 & 1.36 & 4.12 & 276.3 & 0.79 & 0.63 \\
& 400 & & \textbf{1.29} & 4.28 & 253.3 & \textbf{0.79} & 0.64 & \textbf{1.11} & \textbf{4.03} & 297.4 & 0.79 & 0.65 \\
& 640 & & \textbf{1.29} & 4.22 & \textbf{260.4} & \textbf{0.79} & 0.65 & 1.15 & 4.22 & 303.8 & {0.79} & 0.65 \\
\midrule
\multicolumn{13}{l}{\textcolor{figgreen}{\textit{\textbf{Vanilla Latent Diffusion (without foundation-model alignment)}}}} \\
\midrule
DiT~\citep{peebles2023scalable}  & 1400 & 675M & 9.62 & 6.85 & 121.5 & {0.67} & \textbf{0.67} & 2.27 & 4.60 & 278.2 & \textbf{0.83} & 0.57 \\
MaskDiT~\citep{zheng2024fasttrainingdiffusionmodels} & 1600 & 675M & 5.69 & 10.34 & 177.9 & 0.74 & 0.60 & 2.28 & 5.67 & 276.6 & 0.80 & 0.61 \\
SiT~\citep{ma2024sit} & 1400 & 675M & 8.61 & 6.32 & 131.7 & 0.68 & \textbf{0.67} & 2.06 & 4.50 & 270.3 & 0.82 & 0.59 \\
TREAD~\citep{krause2025tread} & 740 & 675M & - & - & - & - & - & 1.69 & 4.73 & 292.7 & 0.81 & 0.63 \\
MDTv2~\citep{gao2023mdtv2} & 1080 & 675M & - & - & - & - & - & 1.58 & 4.52 & \textbf{314.7} & 0.79 & \textbf{0.65} \\
\midrule
\multirow{3}{*}{\textcolor{figpurple}{\textbf{DiT-XL (DM-VAE, \texttt{f16d32})}}} & 64 & \multirow{3}{*}{675M} & 2.62 & 4.62 & 190.1 & 0.78 & 0.60 & 1.67 & 4.12 & 253.4 & 0.80 & 0.62 \\
& 160 & & 1.61 & 4.23 & 222.8 & 0.79 & 0.63 & 1.29 & \textbf{4.01} & 264.3 & 0.80 & 0.64 \\
& 400 & & \textbf{1.25} & \textbf{4.16} & \textbf{250.8} & {0.79} & 0.65 & \textbf{1.17} & {4.02} & 283.0 & 0.80 & 0.64 \\
\midrule
\multirow{3}{*}{\textcolor{figpurple}{\textbf{DiT-XL (DM-VAE, \texttt{f16d64})}}} & 64 & \multirow{3}{*}{675M} & 3.01 & 4.61 & 178.4 & 0.79 & 0.58 & 2.10 & 4.24 & 224.6 & 0.81 & 0.59 \\
& 160 & & 1.71 & 4.32 & 218.1 & \textbf{0.80} & 0.62 & 1.43 & 4.14 & 252.1 & 0.81 & 0.62 \\
& 400 & & {1.28} & \textbf{4.16} & {245.7} & 0.79 & 0.64 & {1.20} & {4.10} & 274.8 & 0.80 & \textbf{0.65} \\
\bottomrule
\end{tabular}
}
\vspace{-1em}
\caption{\textbf{Quantitative comparison of state-of-the-art generative models on ImageNet-1K ($256 \times 256$).} We benchmark our method against vanilla latent diffusion models and foundation-model-aligned LDMs across different training epochs and model capacities. Our method establishes new state-of-the-art results with faster convergence in both foundation-model-free and foundation-model-assisted regimes. Note that our results in the ``w/o CFG'' column are evaluated with single-pass $\text{IG}_{TS}$.}
\vspace{-1em}
\label{tab:sota}
\end{table*}

\begin{table}[t]
\centering
\setlength{\tabcolsep}{3pt}
\scriptsize
\resizebox{\columnwidth}{!}{%
\begin{tabular}{l | ll | ccccc}
\toprule
\textbf{VAE} & \textbf{Align Target} & \textbf{Guidance} & \textbf{FID$_{50K}$}$\downarrow$ & \textbf{sFID}$\downarrow$ & \textbf{IS}$\uparrow$ & \textbf{Prec.}$\uparrow$ & \textbf{Rec.}$\uparrow$ \\
\midrule
SD-VAE~\citep{rombach2022high} & None & None & 25.37 & 5.47 & 52.22 & 0.57 & 0.64 \\
\midrule
\multirow{5}{*}{\textcolor{figpurple}{DM-VAE}} & None & None & 19.03 & 5.47 & 69.06 & 0.61 & \textbf{0.64} \\
& HB ($\bm{z}$) & None & 19.52 & 5.28 & 67.03 & 0.61 & 0.63 \\
& HB ($\bm{z}$) & IG~\citep{zhou2026guiding} & 10.98 & \textbf{4.08} & 91.40 & 0.70 & 0.61 \\
& LB ($\bm{z}_{\text{low}}$) & None & 18.59 & 5.32 & 69.77 & 0.62 & 0.63 \\
& LB ($\bm{z}_{\text{low}}$) & IG$_{TS}$ & \textbf{9.94} & 4.33 & \textbf{97.55} & \textbf{0.71} & 0.60 \\
\bottomrule
\end{tabular}%
}
\vspace{-0.5em}
\caption{\textbf{Ablation study} on DM-VAE, representation alignment, and internal guidance strategy at 80 epochs. Results are obtained under the \texttt{f16d32} setting using \textbf{DiT-B}~\citep{peebles2023scalable} without CFG. HB/LB denote the high/low-dimensional branch: IG~\citep{zhou2026guiding} requires full-channel ($C$-dim) HB predictions, whereas LB alignment predicts only the $C_{\text{low}}$-dim skeleton and thus pairs with $\text{IG}_{TS}$.}
\label{tab:ablation_ig_tra}
\end{table}

\subsection{Ablation Study}
\label{sec:ablation}

We conduct ablation studies on DiT-B at 80 training epochs (without CFG) in Table~\ref{tab:ablation_ig_tra} to isolate the effectiveness of DM-VAE, topological representation alignment (TRA), and internal guidance with topological skeleton ($\text{IG}_{TS}$).

\noindent\textbf{DM-VAE and Low-Dimensional Alignment in TRA.} 
Without any alignment or guidance, replacing SD-VAE with DM-VAE slashes $\text{FID}_{50K}$ from $25.37$ to $19.03$ and boosts IS from $52.22$ to $69.06$, confirming that our dual-manifold regularization yields a much more generation-friendly latent space. On top of DM-VAE, aligning intermediate features $\bm{h}_t$ with HB ($\bm{z}$) degrades $\text{FID}_{50K}$ to $19.52$, since regressing high-frequency details disrupts global structural learning. In contrast, supervising $\bm{h}_t$ with the skeleton LB ($\bm{z}_{\text{low}}$) further reduces $\text{FID}_{50K}$ to $18.59$, providing a clean structural prior for velocity learning.

\noindent\textbf{$\text{IG}_{TS}$ and Standard IG.} 
We further compare standard internal guidance (IG)~\citep{zhou2026guiding} against our $\text{IG}_{TS}$. While standard IG on HB ($\bm{z}$) lowers $\text{FID}_{50K}$ to $10.98$, extrapolating unaligned residuals across all channels amplifies velocity errors and limits semantic quality ($91.40$ IS). Selectively guiding only the smooth skeleton LB ($\bm{z}_{\text{low}}$) while leaving residual details unperturbed achieves the best $\text{FID}_{50K}$ of \textbf{9.94}, IS of \textbf{97.55}, and Precision of \textbf{0.71}.

\section{Conclusion}
\label{sec:conclusion}
In this paper, we identify that the off-manifold problem in high-dimensional VAEs stems from unconstrained channel expansion, whereas relying on external foundation models to regularize the manifold compromises out-of-distribution generalization. To resolve this dilemma internally, we propose the DM-VAE framework, which regularizes the high-dimensional latent space by anchoring it to a smooth low-dimensional topological skeleton. Building upon this, we introduce topological representation alignment (TRA) to provide explicit structural supervision for intermediate features during generative training, and internal guidance with topological skeleton ($\text{IG}_{TS}$) to guide structural components while keeping high-frequency details intact within a single forward pass at inference. Extensive experiments show that our framework reconciles in-domain and OOD reconstruction with fast generative convergence, achieving a $0.12$ rFID and $1.28$/$1.20$ $\text{FID}_{50K}$ (w/o and w/ CFG) at \texttt{f16d64}, $1.25$/$1.17$ at \texttt{f16d32} without external priors, and $1.11$ when orthogonally combined with DINOv2.


\clearpage

\section*{LLM Usage Statement}
LLMs (Gemini) were used solely at the \textit{Assist} level to polish author-written text in the Abstract and Introduction. No LLMs or AI tools were used to generate research ideas or conclusions, compile related work or citations, create figures, or produce data, nor was the manuscript submitted to any LLM for reviewer-style feedback. The authors have verified all content and take full responsibility for the entire submission.
{
    \small
    \bibliographystyle{ieeenat_fullname}
    \bibliography{main}
}

\clearpage
\maketitlesupplementary

\setcounter{section}{0}
\renewcommand{\thesection}{\Alph{section}}

\section{Detailed Off-Manifold Analysis}
\label{app:off_manifold}

We derive the exponential reconstruction distortion bound in Section~\ref{sec:reconstruction}. Under the manifold assumption~\citep{chapelle2009semi,li2026back}, natural images $\bm{x} \in \mathcal{X}$ lie on an intrinsically $d$-dimensional manifold $\mathcal{M}_X$. Let $\bm{u} \in \mathbb{R}^d$ denote the local intrinsic coordinates on $\mathcal{M}_X$ with finite differential entropy $h(\bm{u})$, and let $\hat{\bm{u}} \in \mathbb{R}^d$ denote the intrinsic coordinates of the nearest-point projection of $\hat{\bm{x}}$ onto $\mathcal{M}_X$. Under a locally bi-Lipschitz embedding, the pixel distortion $\varepsilon_{\text{recon}} = \mathbb{E}[\|\bm{x} - \hat{\bm{x}}\|_2^2]$ satisfies $\varepsilon_{\text{recon}} \ge \mu \, \mathbb{E}[\|\bm{u} - \hat{\bm{u}}\|_2^2]$ for a constant $\mu > 0$. In the VAE Markov chain $\bm{u} \leftrightarrow \bm{x} \to \bm{z} \to \hat{\bm{x}} \to \hat{\bm{u}}$, the data processing inequality and the Shannon lower bound~\citep{cover1991elements} yield:
\begin{equation}
\label{eq:app_rd}
\begin{aligned}
    I(\bm{x}; \bm{z}) &\ge I(\bm{u}; \hat{\bm{u}}) = h(\bm{u}) - h(\bm{u} - \hat{\bm{u}} \mid \hat{\bm{u}}) \\
    &\ge h(\bm{u}) - h(\bm{u} - \hat{\bm{u}}) \ge h(\bm{u}) - \frac{d}{2} \log \left( \frac{2\pi e \, \varepsilon_{\text{recon}}}{d \mu} \right) \\
    &= \frac{d}{2} \log \left( \frac{\alpha}{\varepsilon_{\text{recon}}} \right),
\end{aligned}
\end{equation}
where the third inequality follows from the maximum entropy of a $d$-dimensional Gaussian with average coordinate variance $\varepsilon_{\text{recon}}/(d\mu)$, and $\alpha = \frac{d\mu}{2\pi e}\exp(\frac{2}{d}h(\bm{u})) > 0$ is a constant determined by the intrinsic dimension $d$ and intrinsic entropy power of $\mathcal{M}_X$. Conversely, in a continuous VAE with encoder posterior $q_{\mathcal{E}}(\bm{z}\mid\bm{x})$ and Gaussian prior $p(\bm{z}) = \mathcal{N}(\bm{0}, \mathbf{I})$ on $\mathbb{R}^{C \times H \times W}$, the variational information bottleneck bound~\citep{alemi2016deep,alemi2018fixing} upper-bounds the mutual information by the KL divergence $\mathcal{L}_{\text{KL}}$:
\begin{equation}
\label{eq:app_cap}
    I(\bm{x}; \bm{z}) \le \mathbb{E}_{p(\bm{x})}\big[ D_{\text{KL}}\big(q_{\mathcal{E}}(\bm{z}\mid\bm{x}) \,\|\, p(\bm{z})\big) \big] \le \kappa \cdot C H W,
\end{equation}
where $\kappa > 0$ denotes the average KL channel capacity per scalar latent entry. Combining Equations~\eqref{eq:app_rd} and~\eqref{eq:app_cap} gives $\frac{d}{2} \log(\alpha / \varepsilon_{\text{recon}}) \le \kappa C H W$, which establishes the exponential bound in Section~\ref{sec:reconstruction}:
\begin{equation}
    \varepsilon_{\text{recon}} \ge \alpha \exp(-\gamma \cdot C), \quad \text{where} \;\; \gamma = \frac{2\kappa H W}{d} > 0.
\end{equation}

While increasing $C$ exponentially lowers this distortion floor, embedding the $d$-dimensional $\mathcal{M}_X$ into $\mathbb{R}^{C \times H \times W}$ introduces a codimension of $c = C H W - d$ normal directions. Because $\mathcal{L}_{\text{VAE}}$ (Eq.~\eqref{eq:vae_base}) optimizes pixel and adversarial fidelity without constraining this $c$-dimensional normal space, the encoder scatters substantial high-frequency energy along the normal directions, whose large variance disrupts low-frequency structural velocity learning in flow matching (\cf, Appendix~\ref{app:manifold_geometry}).


\begin{figure*}
    \centering
\includegraphics[width=0.94\linewidth]{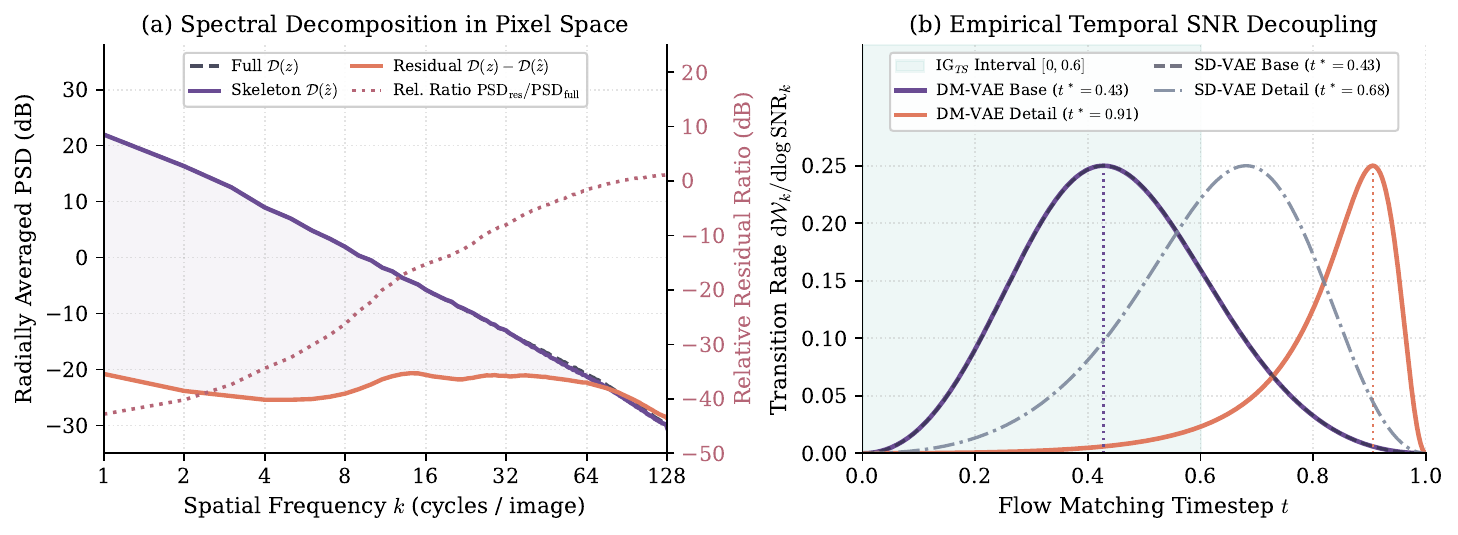}
    \vspace{-0.5em}
    \caption{\textbf{Empirical verification of spectral decomposition and temporal SNR decoupling} on $5\text{K}$ ImageNet validation images (\texttt{f16d64}, $C_{\text{low}}=32$). \textbf{(a)} Radially averaged 2D Fourier PSD in pixel space shows that the skeleton reconstruction $\mathcal{D}(\hat{\bm{z}})$ preserves global low-frequency structures, whereas the detail increment $\mathcal{D}(\bm{z}) - \mathcal{D}(\hat{\bm{z}})$ rises by over $44\text{ dB}$ in relative power ratio ($\mathrm{PSD}_{\mathrm{res}} / \mathrm{PSD}_{\mathrm{full}}$) from $k=1$ to $k=128$. \textbf{(b)} Transition rate curves $\mathrm{d}\mathcal{W}_k/\mathrm{d}\log\mathrm{SNR}_k$ (Eqs.~\eqref{eq:app_wiener}--\eqref{eq:app_t_star}) parameterized by empirical $\sigma_{\text{base}}, \sigma_{\text{detail}}$ on globally unit-variance-normalized latents confirm that while SD-VAE exhibits substantial base--detail overlap ($t_{\text{detail}}^* = 0.68$), DM-VAE pushes the detail transition to $t_{\text{detail}}^* = 0.91$, isolating structural velocity learning within $[0, 0.6]$.}
    \label{fig:app_empirical_snr}
    \vspace{-0.5em}
\end{figure*}

\section{Latent Manifold Geometry Analysis}
\label{app:manifold_geometry}

In this section, we provide a geometric and information-theoretic analysis of how dual-manifold alignment ($\mathcal{L}_{\text{DM}}$) shapes the high-dimensional latent manifold $\mathcal{M}_Z$ into a \textit{generation-friendly tubular structure} and decouples multi-frequency velocity learning in flow matching (FM).

\noindent\textbf{Tubular Geometry of the DM-VAE Latent Space.} 
Recall from Section~\ref{sec:reconstruction} that for any encoded latent $\bm{z} = \mathcal{E}(\bm{x}) \in \mathbb{R}^{C \times H \times W}$, DM-VAE extracts $\bm{z}_{\text{low}} = \mathcal{P}_{\text{down}}(\bm{z}) \in \mathbb{R}^{C_{\text{low}} \times H \times W}$ ($C_{\text{low}} < C$) and lifts it via $\mathcal{P}_{\text{up}}$ to obtain the topological skeleton $\bm{z}^{\text{base}} \triangleq \hat{\bm{z}} = \mathcal{P}_{\text{up}}(\bm{z}_{\text{low}}) \in \mathbb{R}^{C \times H \times W}$. Consistent with Section~\ref{sec:generation}, any latent $\bm{z} \in \mathcal{M}_Z$ can be decomposed into:
\begin{equation}
\label{eq:app_tubular_decomp}
    \bm{z} = \underbrace{\bm{z}^{\text{base}} \, (\equiv \hat{\bm{z}})}_{\text{low-dimensional skeleton}} + \underbrace{\bm{z}^{\text{detail}} \, (\equiv \bm{z} - \hat{\bm{z}})}_{\text{high-frequency residual (details)}},
\end{equation}
where $\bm{z}^{\text{base}}$ lies on the smooth base manifold $\mathcal{M}_{\text{base}} \subset \text{Im}(\mathcal{P}_{\text{up}})$ (of subspace dimension $D_{\text{base}} = C_{\text{low}} H W$) supervised solely by $\mathcal{L}_{\text{rec}}(\bm{x}, \hat{\bm{x}}_{\text{low}})$, and $\bm{z}^{\text{detail}}$ captures fine-grained high-frequency residuals. By the Eckart--Young--Mirsky theorem, minimizing $\mathcal{L}_{\text{DM}} = \mathbb{E}_{\bm{x}}[\|\bm{z} - \mathcal{P}_{\text{up}}\mathcal{P}_{\text{down}}\bm{z}\|_2^2]$ over $(\mathcal{P}_{\text{up}}, \mathcal{P}_{\text{down}})$ naturally favors the rank-$C_{\text{low}}$ orthogonal projector $\mathcal{P} = \mathcal{P}^\top = \mathcal{P}^2$ onto the principal subspace $\text{Im}(\mathcal{P}_{\text{up}})$. Since $\mathcal{P}_{\text{down}}$ is simultaneously supervised by $\mathcal{L}_{\text{rec}}(\bm{x}, \hat{\bm{x}}_{\text{low}})$, in practice we impose a lightweight orthogonality penalty on the $1\times 1$ projection weights ($\mathcal{P}_{\text{down}}\mathcal{P}_{\text{up}} \approx \mathbf{I}, \mathcal{P}_{\text{up}} \approx \mathcal{P}_{\text{down}}^\top$) to preserve Euclidean orthogonality between the two subspaces:
\begin{equation}
\label{eq:app_orth}
    \langle \bm{z}^{\text{base}}, \bm{z}^{\text{detail}} \rangle \approx 0, \quad \|\bm{z}\|_2^2 \approx \|\bm{z}^{\text{base}}\|_2^2 + \|\bm{z}^{\text{detail}}\|_2^2.
\end{equation}
Furthermore, $\mathcal{L}_{\text{DM}}$ directly bounds the expected energy of the transverse residual:
{\small
\begin{equation}
    \mathcal{L}_{\text{DM}} = \mathbb{E}_{\bm{x}} \big[ \| \bm{z} - \hat{\bm{z}} \|_2^2 \big] = \mathbb{E}_{\bm{x}} \big[ \| \bm{z}^{\text{detail}} \|_2^2 \big] \le \epsilon^2 \ll \mathbb{E}_{\bm{x}}\big[\|\bm{z}^{\text{base}}\|_2^2\big].
\end{equation}%
}
By Markov's inequality, $\mathbb{P}(\|\bm{z}^{\text{detail}}\|_2 > \delta) \le \epsilon^2 / \delta^2$ for any radius $\delta > \epsilon$. Consequently, up to a negligible tail probability $\eta = \epsilon^2/\delta^2$, the probability mass of $\mathcal{M}_Z$ is concentrated within a \textit{tubular neighborhood} of radius $\delta$ around $\mathcal{M}_{\text{base}}$:
{\small
\begin{equation}
    \mathcal{T}_{\delta}(\mathcal{M}_{\text{base}}) \triangleq \big\{ \bm{z}^{\text{base}} + \bm{z}^{\text{detail}} \mid \bm{z}^{\text{base}} \in \mathcal{M}_{\text{base}},\, \|\bm{z}^{\text{detail}}\|_2 \le \delta \big\}.
\end{equation}%
}



This tubular concentration provides two key benefits for generation: \textcolor{figpurple}{\textbf{1)}} {\textit{Primary-to-residual velocity learning:}} Since $\mathbb{E}[\|\bm{z}_1^{\text{base}}\|_2^2] \gg \mathbb{E}[\|\bm{z}_1^{\text{detail}}\|_2^2]$, $\mathcal{L}_{\text{FM}}$ naturally prioritizes transporting noise toward the smooth skeleton $\mathcal{M}_{\text{base}}$ before refining local residuals. \textcolor{figpurple}{\textbf{2)}} {\textit{Bounded off-manifold drift:}} Even if the learned velocity $\bm{v}_\theta(\bm{z}_t, t)$ incurs errors along the high-frequency residual directions, the small tubular radius ensures that generated latents remain tightly anchored around the valid skeleton $\mathcal{M}_{\text{base}}$.

\noindent\textbf{Temporal SNR Decoupling in FM Interpolation.} 
Consider the interpolation $\bm{z}_t = t \bm{z}_1 + (1 - t) \bm{z}_0$ ($t \in [0, 1], \bm{z}_0 \sim \mathcal{N}(\bm{0}, \mathbf{I})$). Projecting $\bm{z}_t$ via the orthogonal projectors $\mathcal{P}$ and $\mathbf{I} - \mathcal{P}$ onto the skeleton subspace (of dimension $D_{\text{base}} = C_{\text{low}} H W$) and residual subspace (of dimension $D_{\text{detail}} = (C - C_{\text{low}}) H W$) yields two decoupled processes:
\begin{equation}
    \bm{z}_t^{\text{base}} = t \bm{z}_1^{\text{base}} + (1 - t) \bm{z}_0^{\text{base}}, \quad \bm{z}_t^{\text{detail}} = t \bm{z}_1^{\text{detail}} + (1 - t) \bm{z}_0^{\text{detail}}.
\end{equation}
Within each subspace's orthonormal basis, the projected standard Gaussian noises $\bm{z}_0^{\text{base}}$ and $\bm{z}_0^{\text{detail}}$ are independent and retain unit variance per active subspace dimension ($\frac{1}{D_k}\mathbb{E}[\|\bm{z}_0^k\|_2^2] = 1$ for $k \in \{\text{base}, \text{detail}\}$). Under a zero-mean isotropic Gaussian approximation $\bm{z}_1^k \sim \mathcal{N}(\bm{0}, \sigma_k^2 \mathbf{I}_{D_k})$ (encouraged by the standard normal KL prior) with per-subspace-dimension signal variance $\sigma_k^2 = \frac{1}{D_k}\mathbb{E}[\|\bm{z}_1^k\|_2^2]$, the instantaneous subspace signal-to-noise ratio (SNR) at timestep $t$ is:
\begin{equation}
    \text{SNR}_k(t) = \frac{t^2 \sigma_k^2}{(1 - t)^2}, \quad k \in \{\text{base}, \text{detail}\}.
\end{equation}
Minimizing the $L_2$ loss $\mathcal{L}_{\text{FM}}$ yields the optimal velocity $\bm{v}^*(\bm{z}_t, t) = \mathbb{E}[\bm{z}_1 - \bm{z}_0 \mid \bm{z}_t] = \frac{\mathbb{E}[\bm{z}_1 \mid \bm{z}_t] - \bm{z}_t}{1 - t}$, where the minimum mean-squared error (MMSE) denoiser $\mathbb{E}[\bm{z}_1^k \mid \bm{z}_t^k]$ in each subspace takes the closed-form Wiener filter expression:
\begin{equation}
\label{eq:app_wiener}
    \mathbb{E}[\bm{z}_1^k \mid \bm{z}_t^k] = \frac{1}{t} \underbrace{\frac{\text{SNR}_k(t)}{1 + \text{SNR}_k(t)}}_{\mathcal{W}_k(t)} \bm{z}_t^k, \quad k \in \{\text{base}, \text{detail}\}.
\end{equation}
When $\text{SNR}_k(t) \ll 1$, the Wiener gain $\mathcal{W}_k(t) \approx 0$ (the signal $\bm{z}_1^k$ is submerged in noise and unrecoverable); when $\text{SNR}_k(t) \gg 1$, $\mathcal{W}_k(t) \approx 1$ (trivial linear recovery). The rate of information transition with respect to log-SNR, $\frac{\mathrm{d}\mathcal{W}_k}{\mathrm{d}\log\text{SNR}_k} = \frac{\text{SNR}_k}{(1 + \text{SNR}_k)^2}$, attains its unique global maximum at $\text{SNR}_k(t_k^*) = 1$~\citep{esser2024scaling}, yielding the critical transition timestep:
\begin{equation}
\label{eq:app_t_star}
    \frac{t_k^*}{1 - t_k^*} = \frac{1}{\sigma_k} \;\Longrightarrow\; t_k^* = \frac{1}{1 + \sigma_k}.
\end{equation}
Equation~\eqref{eq:app_t_star} reveals a key distinction between unconstrained VAEs and DM-VAE: \textcolor{figpurple}{\textbf{1)}} {\textit{Spectral interference in unconstrained VAEs:}} In standard high-reconstruction VAEs (\eg, SD-VAE~\citep{rombach2022high}), unregularized high-frequency details carry per-mode variance comparable to low-frequency structures ($\sigma_{\text{detail}} \approx \sigma_{\text{base}}$), causing their transition windows to overlap ($t_{\text{detail}}^* \approx t_{\text{base}}^*$). Because high-frequency targets exhibit large sample-to-sample variance, regressing them simultaneously at intermediate timesteps injects severe gradient variance that disrupts global structural learning. \textcolor{figpurple}{\textbf{2)}} {\textit{Natural temporal decoupling in DM-VAE:}} By suppressing transverse residual energy via $\mathcal{L}_{\text{DM}}$ ($\sigma_{\text{detail}} \ll \sigma_{\text{base}}$), Equation~\eqref{eq:app_t_star} strictly separates the two transition windows:
\begin{equation}
    t_{\text{base}}^* = \frac{1}{1 + \sigma_{\text{base}}} \;<\; t_{\text{detail}}^* = \frac{1}{1 + \sigma_{\text{detail}}} \;\longrightarrow\; 1.
\end{equation}
At intermediate timesteps $t \approx t_{\text{base}}^*$, $\text{SNR}_{\text{base}}(t) \approx 1$ while $\text{SNR}_{\text{detail}}(t) \ll 1$ ($\mathcal{W}_{\text{detail}}(t) \approx 0$), allowing the network to learn the low-variance velocity toward the global skeleton $\bm{z}_1^{\text{base}}$ without high-frequency interference, before resolving fine textures near $t \to 1$.

\begin{figure}
    \centering
    \includegraphics[width=0.97\linewidth]{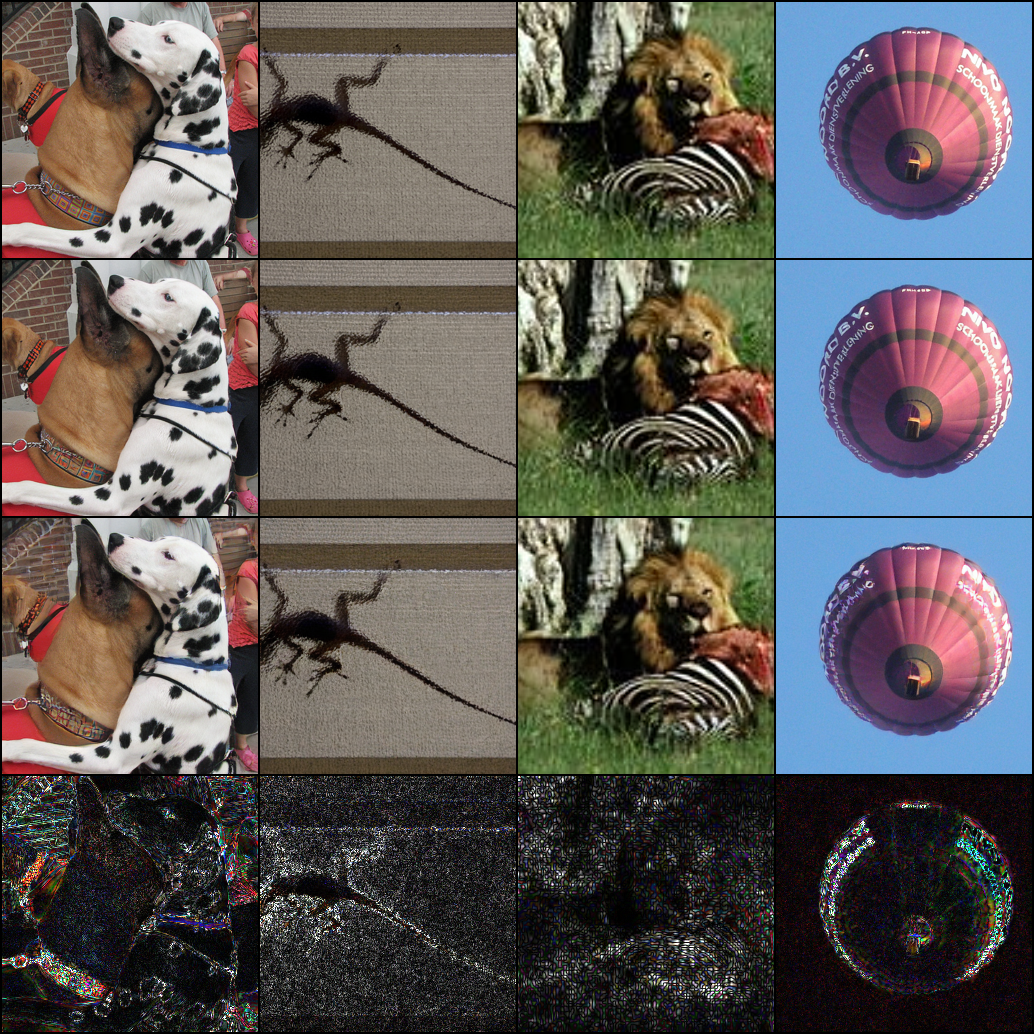}
    \vspace{-0.5em}
    \caption{\textbf{Qualitative visualization of latent subspace decomposition} on ImageNet validation samples using DM-VAE (\texttt{f16d64}, $C_{\text{low}}=32$). From top to bottom: input image $\bm{x}$, full reconstruction $\mathcal{D}(\bm{z})$, decoded skeleton $\mathcal{D}(\hat{\bm{z}})$, and residual $|\mathcal{D}(\bm{z}) - \mathcal{D}(\hat{\bm{z}})|$. While $\mathcal{D}(\hat{\bm{z}})$ preserves global structure and color layout, the residual isolates fine-grained edges and high-frequency textures.}
\label{fig:decomposition_visual_samples}
\end{figure}

\noindent\textbf{Empirical Verification.} 
Figures~\ref{fig:app_empirical_snr} and~\ref{fig:decomposition_visual_samples} empirically validate our geometric and SNR analysis on $5\text{K}$ ImageNet validation images under \texttt{f16d64} ($C_{\text{low}}=32$). First, encoded latents exhibit near-exact subspace orthogonality with $|\cos(\bm{z}^{\text{base}}, \bm{z}^{\text{detail}})| = \textbf{0.0047}$, corroborating Eq.~\eqref{eq:app_orth}. In pixel space (Figure~\ref{fig:app_empirical_snr}(a)), $\mathcal{D}(\hat{\bm{z}})$ closely matches $\mathcal{D}(\bm{z})$ across global low frequencies ($k \le 8$, with $\mathrm{PSD}_{\mathrm{res}} / \mathrm{PSD}_{\mathrm{full}} < -27\text{ dB}$), while the detail increment $\mathcal{D}(\bm{z}) - \mathcal{D}(\hat{\bm{z}})$ concentrates on intra-patch high-frequency textures ($k \ge 16$). Second, in latent space (\cf, Figure~\ref{fig:app_empirical_snr}(b)), even when granting SD-VAE its optimal post-hoc rank-$32$ PCA projector (the theoretical lower bound on $\sigma_{\text{detail}}$), its detail transition window ($t_{\text{detail}}^* = 0.68$) heavily overlaps with the base window ($t_{\text{base}}^* = 0.43$) around intermediate timesteps. Conversely, DM-VAE suppresses $\sigma_{\text{detail}}$ to push $t_{\text{detail}}^*$ to $0.91$ ($\Delta t^* = 0.48$ vs.\ $0.25$), keeping detail transition rates near zero throughout $[0, 0.6]$.

\section{Detailed Training Settings}
\label{app:training_settings}
\noindent\textbf{DM-VAE Training.} 
We train DM-VAE on the ImageNet-1K~\citep{deng2009imagenet} training set at $256 \times 256$ resolution (with random cropping and horizontal flipping), following the architectural and optimization conventions of SD-VAE~\citep{rombach2022high} and VA-VAE~\citep{yao2025reconstruction}. 
\textcolor{figpurple}{\textbf{1)}} {\textit{Architecture:}} The shared encoder $\mathcal{E}$ and main decoder $\mathcal{D}$ adopt a 5-stage convolutional backbone with channel multipliers $(1, 1, 2, 2, 4)$ (spatial downsampling factor $f=16$), base channel width $128$, $2$ residual blocks per stage, and self-attention at the $16 \times 16$ latent resolution. To ensure efficiency and structural inductive bias, the auxiliary lightweight decoder $\mathcal{D}_{\text{low}}$ halves the base channel width to $64$, uses $1$ residual block per stage, and removes all self-attention layers; $\mathcal{D}_{\text{low}}$ is used only during training and discarded at inference. The linear projections $\mathcal{P}_{\text{down}}$ and $\mathcal{P}_{\text{up}}$ are implemented as bias-free $1 \times 1$ convolutions with orthogonal initialization ($\mathbf{W}_{\text{down}} = \mathbf{W}_{\text{up}}^\top$). 
\textcolor{figpurple}{\textbf{2)}} {\textit{Objectives and adaptive weighting:}} Both $\mathcal{L}_{\text{rec}}(\bm{x}, \hat{\bm{x}})$ and $\mathcal{L}_{\text{rec}}(\bm{x}, \hat{\bm{x}}_{\text{low}})$ sum pixel-level $L_1$ and perceptual LPIPS~\citep{zhang2018unreasonable} losses with unit weights. We set the KL penalty weight $\beta = 10^{-6}$ and the low-dimensional branch weight $\lambda_{\text{low}} = 0.2$. Adversarial training ($\mathcal{L}_{\text{adv}}$) is applied exclusively to the main reconstruction $\hat{\bm{x}}$ via a 3-layer PatchGAN discriminator~\citep{esser2021taming} with hinge loss after $1{,}000$ warmup steps, balanced adaptively by the gradient-norm ratio at the last convolutional layer of $\mathcal{D}$ (with base weight $0.5$). For dual-manifold alignment, $\mathcal{L}_{\text{DM}}$ is computed on the posterior means $\bm{\mu}$ and $\hat{\bm{\mu}} = \mathcal{P}_{\text{up}}(\mathcal{P}_{\text{down}}(\bm{\mu}))$ using the symmetric stop-gradient ($\text{sg}[\cdot]$) formulation in Section~\ref{sec:reconstruction}. Its effective coefficient $\lambda_{\text{DM}} = 0.1 \,w_{\text{DM}}$ is dynamically balanced every $20$ steps via an exponential moving average (EMA, decay $0.8$) of the gradient-norm ratio at the shared pre-latent layer $\theta_{\mathcal{E}}^{\text{last}}$:
\begin{equation}
    w_{\text{DM}} \leftarrow 0.8 \, w_{\text{DM}} + 0.2 \, \frac{\|\nabla_{\theta_{\mathcal{E}}^{\text{last}}} \mathcal{L}_{\text{rec}}(\bm{x}, \hat{\bm{x}})\|_2}{\|\nabla_{\theta_{\mathcal{E}}^{\text{last}}} \|\bm{\mu} - \text{sg}[\hat{\bm{\mu}}]\|_2^2\|_2 + 10^{-2}}.
\end{equation}
In addition, let $\mathbf{W}_{\text{down}} \in \mathbb{R}^{C_{\text{low}} \times C}$ and $\mathbf{W}_{\text{up}} \in \mathbb{R}^{C \times C_{\text{low}}}$ denote the reshaped $1 \times 1$ kernel matrices of $\mathcal{P}_{\text{down}}$ and $\mathcal{P}_{\text{up}}$. Consistent with Appendix~\ref{app:manifold_geometry}, we include an orthogonality regularizer $\lambda_{\text{ortho}} \mathcal{L}_{\text{ortho}}$ ($\lambda_{\text{ortho}} = 10.0$) purely as a safeguard against degenerate rescaling or basis drift; given our orthogonal initialization, $\mathcal{L}_{\text{ortho}}$ remains near zero throughout training and \textit{exerts negligible influence} on optimization:
\begin{equation}
\begin{aligned}
    \mathcal{L}_{\text{ortho}} &= \text{MSE}(\mathbf{W}_{\text{down}}\mathbf{W}_{\text{down}}^\top, \mathbf{I}) + \text{MSE}(\mathbf{W}_{\text{up}}^\top\mathbf{W}_{\text{up}}, \mathbf{I}) \\
    &\quad + 0.5 \, \text{MSE}(\mathbf{W}_{\text{down}}, \mathbf{W}_{\text{up}}^\top).
\end{aligned}
\end{equation}
\textcolor{figpurple}{\textbf{3)}} {\textit{Optimization:}} Both DM-VAE and the discriminator are trained within $50$ epochs using Adam ($\text{lr} = 1 \times 10^{-4}$, $\beta_1 = 0.5, \beta_2 = 0.9$) in \texttt{bfloat16} mixed precision with a global batch size of $256$, gradient norm clipping at $1.0$, and an EMA decay of $0.9999$.

We omit adversarial loss on the low-dimensional branch ($\hat{\bm{x}}_{\text{low}}$) for both \textit{training efficiency} and \textit{optimization stability}. First, incorporating an auxiliary adversarial objective noticeably slows down DM-VAE training due to additional discriminator passes and adaptive gradient-norm evaluations. Second, adversarial training is inherently unstable under a strict capacity bottleneck: in our preliminary experiments, applying a GAN loss to $\hat{\bm{x}}_{\text{low}}$ consistently triggered late-stage adversarial collapse, as the discriminator easily overpowered the capacity-limited lightweight decoder $\mathcal{D}_{\text{low}}$ and broke the minimax equilibrium.

\noindent\textbf{DiT Training.} 
We conduct class-conditional generative training on ImageNet-1K ($256 \times 256$)~\citep{deng2009imagenet} strictly adhering to the official settings of LightningDiT~\citep{yao2025reconstruction} and IG~\citep{zhou2026guiding}. For topological representation alignment ($\mathcal{L}_{\text{TRA}}$), the intermediate features $\bm{h}_t$ are extracted from the same default transformer layer as in IG~\citep{zhou2026guiding} and mapped to $\mathbb{R}^{C_{\text{low}} \times H \times W}$ via a lightweight MLP projection head $\mathcal{P}_{\text{align}}$. For fair evaluation at inference, both our methods and all reproduced baselines adopt an identical $250$-step Euler solver with uniform class-balanced sampling across all classes.

\section{More Results}
\label{app:more_results}

\begin{table}[t]
\centering
\setlength{\tabcolsep}{1pt}
\scriptsize
\setlength{\tabcolsep}{4.5pt}
\resizebox{\columnwidth}{!}{%
\begin{tabular}{l | ccccc}
\toprule
\textbf{Methods} & \textbf{FID$_{50K}$}$\downarrow$ & \textbf{sFID}$\downarrow$ & \textbf{IS}$\uparrow$ & \textbf{Prec.}$\uparrow$ & \textbf{Rec.}$\uparrow$ \\
\midrule
Improved-SiT-B$^\dagger$~\citep{ma2024sit} & 25.1 & 6.54 & 58.8 & 0.57 & 0.64 \\
Improved-REPA-B$^\dagger$~\citep{yu2024representation} & 19.1 & 6.88 & 76.5 & 0.60 & \textbf{0.66} \\
\textcolor{figpurple}{DDT-B (DM-VAE)} & 15.2 & 6.08 & 84.0 & 0.63 & \textbf{0.66} \\
\textcolor{figpurple}{DDT-B (DM-VAE$_{\textit{Dino}}$)} & 11.0 & 5.47 & 94.8 & 0.70 & 0.62 \\
\quad + \textcolor{figpurple}{IG$_{TS}$} & \textbf{5.1} & \textbf{5.23} & \textbf{136.9} & \textbf{0.77} & 0.57 \\
\midrule
Improved-SiT-XL$^\dagger$~\citep{ma2024sit} & 10.9 & 5.30 & 103.4 & 0.66 & 0.65 \\
Improved-REPA-XL$^\dagger$~\citep{yu2024representation} & 8.1 & 5.34 & 124.9 & 0.68 & 0.67 \\
\textcolor{figpurple}{DDT-XL (DM-VAE)} & 5.9 & 4.75 & 139.8 & 0.70 & \textbf{0.68} \\
\textcolor{figpurple}{DDT-XL (DM-VAE$_{\textit{Dino}}$)} & 4.1 & 4.43 & 157.0 & 0.75 & 0.64 \\
\quad + \textcolor{figpurple}{IG$_{TS}$} & \textbf{2.0} & \textbf{4.35} & \textbf{213.3} & \textbf{0.80} & 0.61 \\
\bottomrule
\end{tabular}%
}
\vspace{-1em}
\caption{\textbf{Generation performance comparison on ImageNet} at 80 generative training epochs using \textbf{DDT}~\citep{wang2026ddt} without CFG. $^\dagger$ denotes baselines improved by SwiGLU~\citep{touvron2023llama}, RoPE~\citep{su2024roformer}, and RMSNorm~\citep{touvron2023llama}. Our DM-VAEs adopt the \texttt{f16d32} configuration. For IG$_{TS}$, we consistently apply a guidance scale of 1.7 across the timestep interval of $[0, 0.6]$.}
\vspace{-0.5em}
\label{tab:ddt}
\end{table}

\noindent\textbf{Quantitative Results on DDT.} 
To verify the generality of our framework beyond standard DiT, we further integrate DM-VAE (\texttt{f16d32}) and $\text{IG}_{TS}$ into the Decoupled Diffusion Transformer (DDT)~\citep{wang2026ddt} across Base (B) and Extra-Large (XL) scales at 80 training epochs without CFG (\cf, Table~\ref{tab:ddt}). 
Using our foundation-model-free DM-VAE without guidance, {DDT (DM-VAE)} substantially outperforms both Improved-SiT~\citep{ma2024sit} ($25.1 \to 15.2$ $\text{FID}_{50K}$ on B; $10.9 \to 5.9$ on XL) and Improved-REPA~\citep{yu2024representation} ($19.1 \to 15.2$ on B; $8.1 \to 5.9$ on XL), while simultaneously improving spatial fidelity ($6.08$/$4.75$ sFID) and diversity ($0.66$/$0.68$ Recall). Incorporating DINOv2 alignment in {DM-VAE}$_{\textit{Dino}}$ further reduces $\text{FID}_{50K}$ to $11.0$ (B) and $4.1$ (XL). When combined with single-pass $\text{IG}_{TS}$, generation quality improves, reaching \textcolor{figpurple}{\textbf{5.1}} $\text{FID}_{50K}$ (\textcolor{figpurple}{\textbf{136.9}} IS) on DDT-B and \textcolor{figpurple}{\textbf{2.0}} $\text{FID}_{50K}$ (\textcolor{figpurple}{\textbf{4.35}} sFID, \textcolor{figpurple}{\textbf{213.3}} IS) on DDT-XL without CFG, confirming that our dual-manifold alignment and skeleton guidance generalize across distinct architectures.

\begin{figure*}
    \centering
    \includegraphics[width=0.96\linewidth]{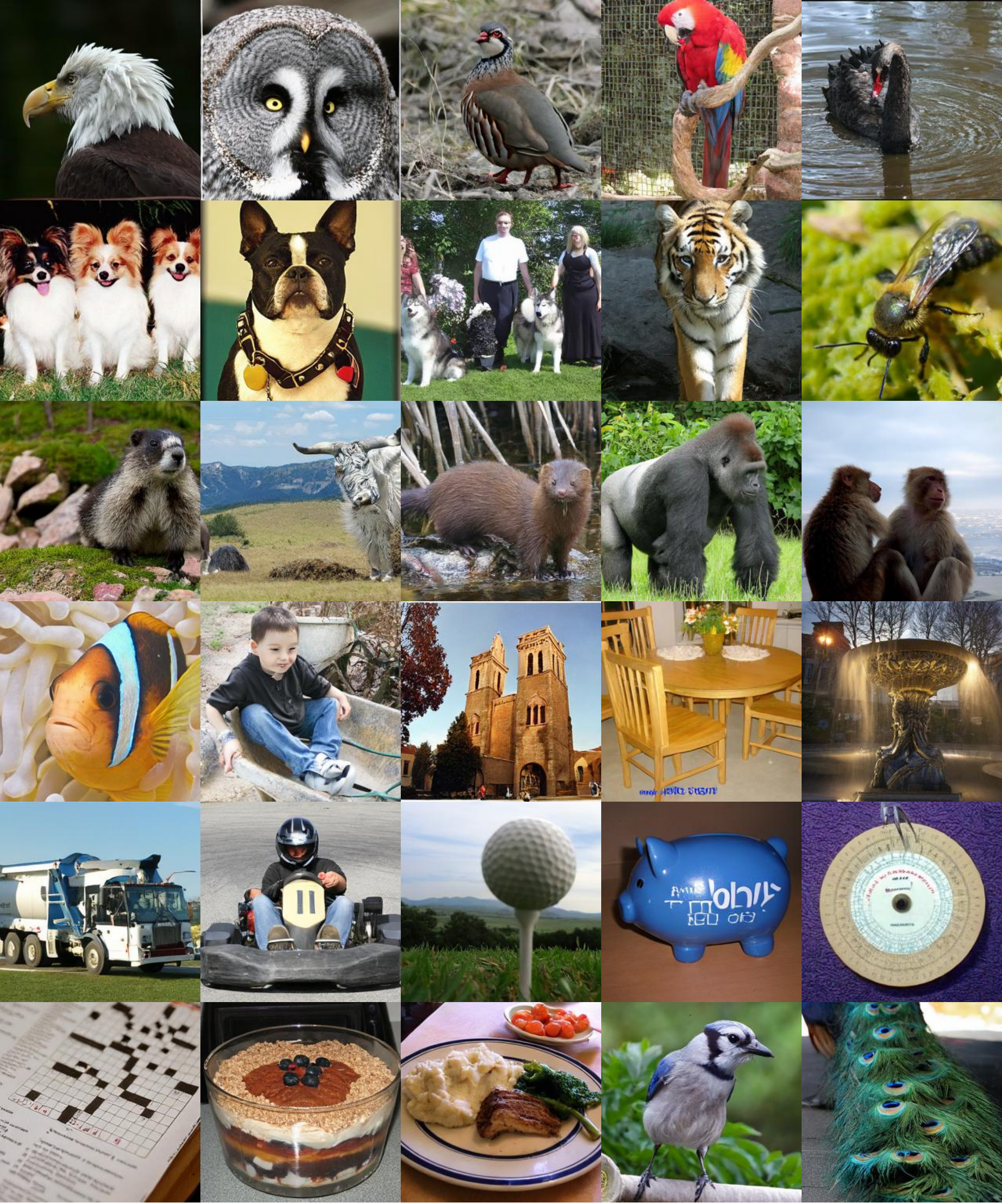}
    \vspace{-0.5em}
    \caption{\textbf{Qualitative generation results} using DiT-XL with our foundation-model-free DM-VAE (\texttt{f16d64}) at \textbf{400 epochs.}}
    \label{fig:samples_1}
    \vspace{-1em}
\end{figure*}

\begin{figure*}
    \centering
    \includegraphics[width=0.96\linewidth]{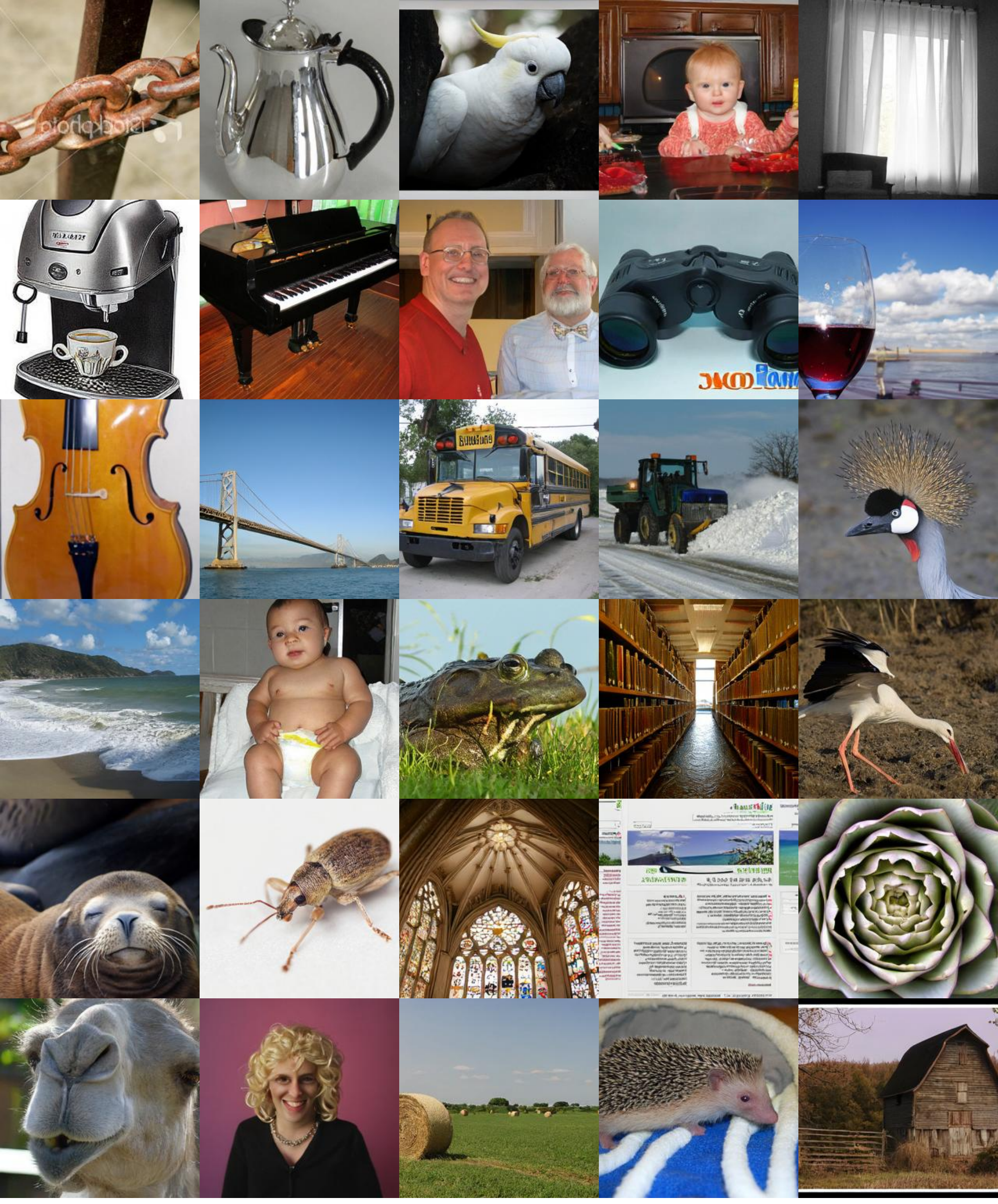}
    \vspace{-0.5em}
    \caption{\textbf{Qualitative generation results} using DiT-XL with our foundation-model-free DM-VAE (\texttt{f16d64}) at \textbf{400 epochs.}}
    \label{fig:samples_2}
    \vspace{-1em}
\end{figure*}

\noindent\textbf{Qualitative Generation Results.} We provide more visual samples in Figures~\ref{fig:samples_1} and~\ref{fig:samples_2} without using CFG.

\noindent\textbf{Impact of Low Dimension $C_{\text{low}}$.}
We further investigate the choice of the low-dimensional bottleneck $C_{\text{low}}$ under the \texttt{f16d64} ($C=64$) configuration. On DiT-XL at 80 training epochs (without CFG or $\text{IG}_{TS}$), setting $C_{\text{low}}=32$ achieves an $FID_{10K}$ of 10.40, whereas further compressing it to $C_{\text{low}}=16$ degrades $FID_{10K}$ to 12.00. This reflects an inherent trade-off in selecting $C_{\text{low}}$, which should be neither overly large nor overly small. An excessively large $C_{\text{low}}$ weakens the information bottleneck, leaving $\bm{z}_{\text{low}}$ itself unconstrained with excessive degrees of freedom and prone to manifold degradation, which in turn compromises its topological regularization on the high-dimensional latent $\bm{z}$. Conversely, an overly small $C_{\text{low}}$ severely restricts skeleton capacity and fails to adequately suppress transverse high-frequency residual energy, widening the tubular neighborhood around lifted $\bm{z}_{\text{low}}$ and limiting its ability to mitigate the off-manifold problem. Balancing skeleton smoothness and residual energy suppression, we set $C_{\text{low}} = C/2$ by default across our main experiments.

\clearpage

\end{document}